%% file: main.tex
\documentclass{article}
\usepackage{arxiv}

\pdfoutput=1
\usepackage[pdftex]{}

\input{preamble}

\usepackage{amsmath, amsthm, amssymb}

\newtheorem{proposition}{Proposition}

\usepackage{caption}

\title{CLIPArTT: Adaptation of CLIP to New Domains at Test Time}

\date{} 					% Or removing it

\author{ Gustavo A. Vargas Hakim\thanks{Equal contribution} \And David Osowiechi \footnotemark[1] \And Mehrdad Noori \And Milad Cheraghalikhani \And Ali Bahri \And Moslem Yazdanpanah \And Ismail Ben Ayed \And Christian Desrosiers}
\date{LIVIA, ÉTS Montréal, Canada \\ International Laboratory on Learning Systems (ILLS), \\ McGILL - ETS - MILA - CNRS - Université Paris-Saclay - CentraleSupélec, Canada \texttt{gustavo-adolfo.vargas-hakim.1@ens.etsmtl.ca, david.osowiechi.1@ens.etsmtl.ca}}

\renewcommand{\shorttitle}{CLIPArTT: Light-weight Adaptation of CLIP to New Domains at Test Time}

\hypersetup{
pdftitle={main},
pdfsubject={cs.ML, cs.AI, cs.CV},
pdfauthor={Gustavo A. Vargas Hakim, David Osowiechi, Mehrdad Noori, Milad Cheraghalikhani, Ismail Ben Ayed, Christian Desrosiers},
pdfkeywords={Test-time Adaptation, Image Classification, Mutual Information, Unsupervised Training},
}

\begin{document}
\maketitle

%%%%%%%%% ABSTRACT
\begin{abstract}
   Pre-trained vision-language models (VLMs), exemplified by CLIP, demonstrate remarkable adaptability across zero-shot classification tasks without additional training. However, their performance diminishes in the presence of domain shifts. In this study, we introduce CLIP Adaptation duRing Test-Time (CLIPArTT), a fully test-time adaptation (TTA) approach for CLIP, which involves automatic text prompts construction during inference for their use as text supervision. Our method employs a unique, minimally invasive text prompt tuning process, wherein multiple predicted classes are aggregated into a single new text prompt, used as \emph{pseudo label} to re-classify inputs in a transductive manner. Additionally, we pioneer the standardization of TTA benchmarks (e.g., TENT) in the realm of VLMs. Our findings demonstrate that, without requiring additional transformations nor new trainable modules, CLIPArTT enhances performance dynamically across non-corrupted datasets such as CIFAR-100, corrupted datasets like CIFAR-100-C and ImageNet-C, alongside synthetic datasets such as VisDA-C. This research underscores the potential for improving VLMs' adaptability through novel test-time strategies, offering insights for robust performance across varied datasets and environments. The code can be found at: \url{https://github.com/dosowiechi/CLIPArTT.git}
\end{abstract}

%%%%%%%%% BODY TEXT
\section{Introduction}
\label{sec:intro}

Combining vision and language modalities for learning, namely a Vision Language model (VLM), has demonstrated an outstanding performance in different vision tasks \cite{clip,align,sam}. Remarkably, these models are surprisingly effective at zero-shot generalization, where a new task can be outside the original scope of the training set, without any additional supervision to fine tune the model. Models such as CLIP \cite{clip} have then been employed in fields as diverse as video recognition \cite{clipvideo}, audio \cite{clipaudio}, and medical imaging \cite{clipmedical}. These advancements point these methods to become critical in future machine learning research and applications.

As in other more traditional deep architectures (e.g., CNNs), CLIP is prone to performance degradation on domains to which it was not originally exposed. Recent research trends suggest that domain adaptation mechanisms can play an important role in deploying CLIP \cite{padclip,tpt}. The challenge, however, is to adapt the model to new domains in real-time and as fast as possible, to maintain its attractive zero-shot capabilities without the need of retraining.

In this paper, CLIP is contextualized in the setting of Test-Time Adaptation, a challenging yet practical scenario of domain adaptation. In realistic scenarios, a model needs to adapt to new data \emph{on-the-fly} to cope with unknown distribution shifts, and without using any class supervision. Although an experimental boilerplate was standardized in recent years, CLIP has not been integrated to it yet. Additionally, we introduce a powerful adaptation technique that achieves \emph{state-of-the-art} performance without a significant computational overhead. Comprehensive studies are performed on multiple datasets containing different types of domain shifts on several levels of severity, resulting in a total of numerous evaluation scenarios. Our main contributions can be summarized as follows:
\begin{itemize}\setlength\itemsep{.1em}
    \item We propose a Test-Time Adaptation method for CLIP, named CLIPArTT, which adapts the VLM by updating normalization-layer parameters. This is achieved by combining multiple classes into a single new text prompt, which is then used as a pseudo-label.
    \item We introduce a new benchmark for Test-Time Adaptation on Vision-Language Models by implementing representative baselines, such as TENT.
    \item Through comprehensive experiments, we subject our CLIPArTT methodology to diverse and challenging Test-Time Adaptation scenarios, each characterized by distinct types of domain shifts. The outcomes of these experiments highlight the performance of our approach when compared against other methodologies addressing similar challenges.
\end{itemize}

%The rest of the paper is organized in the following sections. Section~\ref{sec:related} compiles significant related work to contextualize our method. Section~\ref{sec:methodology} introduces CLIPArTT and explains its functionality. Section~\ref{sec:experiments} details the experimental setting, with results exposed in Section~\ref{sec:results}. Section~\ref{sec:conclusions} concludes with a summary of contributions and potential future work.

\section{Related work}
\label{sec:related}

\mypar{Test-Time Adaptation.} TTA is a particular setting of Domain Adaptation, encompassing two main characteristics: (\emph{a}) adapting a model to a target domain, with inputs coming as unlabeled data streams (i.e., batches), without (\emph{b}) any access to the source domain samples. The former challenge complicates accurately estimating the target domain's distribution, while the latter impedes solving the problem by directly comparing domains' characteristics such as statistics. Despite the challenging nature of the problem, the field has gained important momentum in recent years, providing insights on the possibilities and limitations of adapting pre-trained models. 

A key focus in TTA methods is adapting batch normalization layers, which retain important source domain information. PTBN \cite{PTBN} adjusts batch statistics at test time, while TENT \cite{tent} refines affine parameters using entropy minimization on predictions. Entropy minimization, used in various methods \cite{conjugate,memo,ttc,wisdom}, enhances model confidence without label supervision but often depends on image augmentations or large batches. Text-Prompt Tuning (TPT) \cite{tpt} applies entropy minimization to CLIP by learning text prompt adapters, though it is costly due to required augmentations. Test-Time Distribution Normalization (TTDN) \cite{TTDN} normalizes test data to match the training distribution but needs access to source data or approximations. Our method fine-tunes normalization layers, leverages prediction confidence for text supervision, and avoids input augmentations.

Recent techniques have also sought to efficiently adapt CLIP in a gradient-free manner. CALIP~\cite{calip} adapts the visual and text features bidirectionally through a parametric-free attention module, and later combine them to obtain new logits. Although efficient, this method requires the features of the entire test set to perform a hyperparameter search, which limits its generalization to very large datasets. TDA~\cite{tda} dynamically builds a positive and a negative cache to adapt features through the Tip-Adapter strategy~\cite{tip}. This method however requires to find specific hyperparameters for each dataset, as in the original Tip-Adapter approach.

Pseudo-labels have been central to previous TTA methods. SHOT \cite{shot} uses them with cross-entropy to regularize mutual information, optimizing the entire feature encoder. CoTTA \cite{cotta} employs pseudo-labels in a student-teacher model with consistency loss between original and augmented inputs. PAD \cite{pad} enhances pseudo-labels by augmenting inputs and voting on predictions. Instead of simple class pseudo-labels, our CLIPArTT method improves the text supervision to broaden the likelihood of accurately predicting the right class.

Other methods take less conventional approaches to TTA. LAME \cite{lame} refines classifier predictions transductively using feature similarity through the Laplacian. Test-time training (TTT) methods \cite{ttt,ttt++,tttflow,tttmask,clust3} train a sub-branch alongside the main network in an unsupervised manner to update the model, requiring training from scratch on the source domain. Like LAME, our method relates to Laplacian regularization but applies it differently. While LAME refines predictions without altering the model, our method uses it in a test-time adaptation loss to enforce consistency between embeddings of related batch samples. Unlike TTT methods, we do not require additional branches or training from scratch.

\mypar{Conformal Learning.} CLIPArTT is also related to the field of conformal learning, where intervals of confidence for new predictions are derived from previous experience \cite{conformal1}. In a conformal prediction, a given level of certainty is assigned to a set $\mathcal{C}=\{c_{1}, ..., c_{K}\}$ with the $K$ most plausible classes that an input can belong to \cite{conformal2}. We draw inspiration from this concept and build a conformal set of class predictions that can help adapting CLIP towards an accurate top-1 prediction. Our technique, however, stands out by not relying on image transformations such as in \cite{tpt,padclip} to filter out predictions.

\section{Methodology}
\label{sec:methodology}

We start by presenting the vanilla CLIP model for classification and explain how it can be extended to test-time adaptation using entropy minimization. Building on the limitations of this approach, we then introduce our CLIPArTT method that leverages class uncertainty and the relationship between samples in a batch.

\begin{figure*}[t!]
    \centering
    \includegraphics[width=\textwidth]{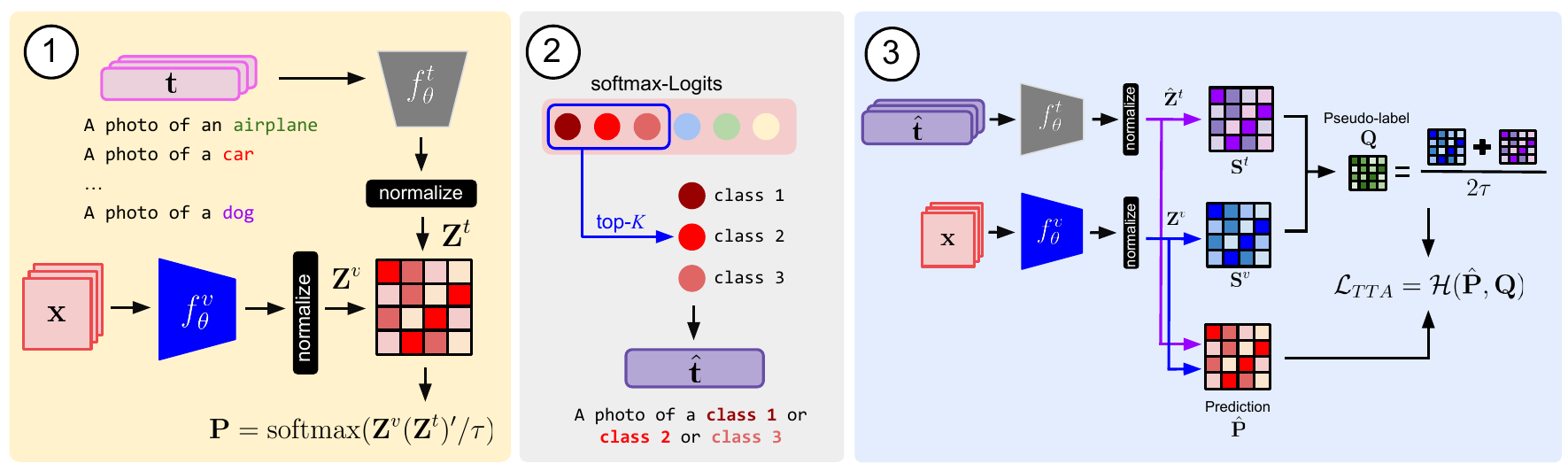}
    \caption{
    % The overall architecture of CLIPArTT. 1) We compute the similarity between the image features and the text features of all the classes. 2) We take the top $K$ classes predicted by the model to make a new text prompts for each image. 3) We use these new features as pseudo labels to compute the cross entropy with the logits.
    CLIPArTT pipeline overview: 1) Computing predictions from Image-Text Similarity, 2) generating a new text prompt by filtering the top-$K$ class predictions, 3) with the new prompts, a pseudo-label $\textbf{Q}$ is obtained by averaging the image-to-image and text-to-text similarity scores, while the prediction $\hat{\textbf{P}}$ is computed as the image-to-text similarity. Cross-entropy is then used as the TTA loss.}
    \label{fig:diagram}
\end{figure*}

\begin{comment}
\begin{figure*}
    \centering
    \includegraphics[width=\textwidth]{Figures/similarity.pdf}
    \caption{Similarity matrices before (a) and after (b) adaptation using $K=5$ classes. There is an increase in the confidence of the logits when using a pseudolabel from the new prompt with the $K$ most confident classes.}
    \label{fig:similarity}
\end{figure*}
\end{comment}

\begin{figure*}
    \centering
    \begin{tabular}{c}
    \includegraphics[width=\textwidth]{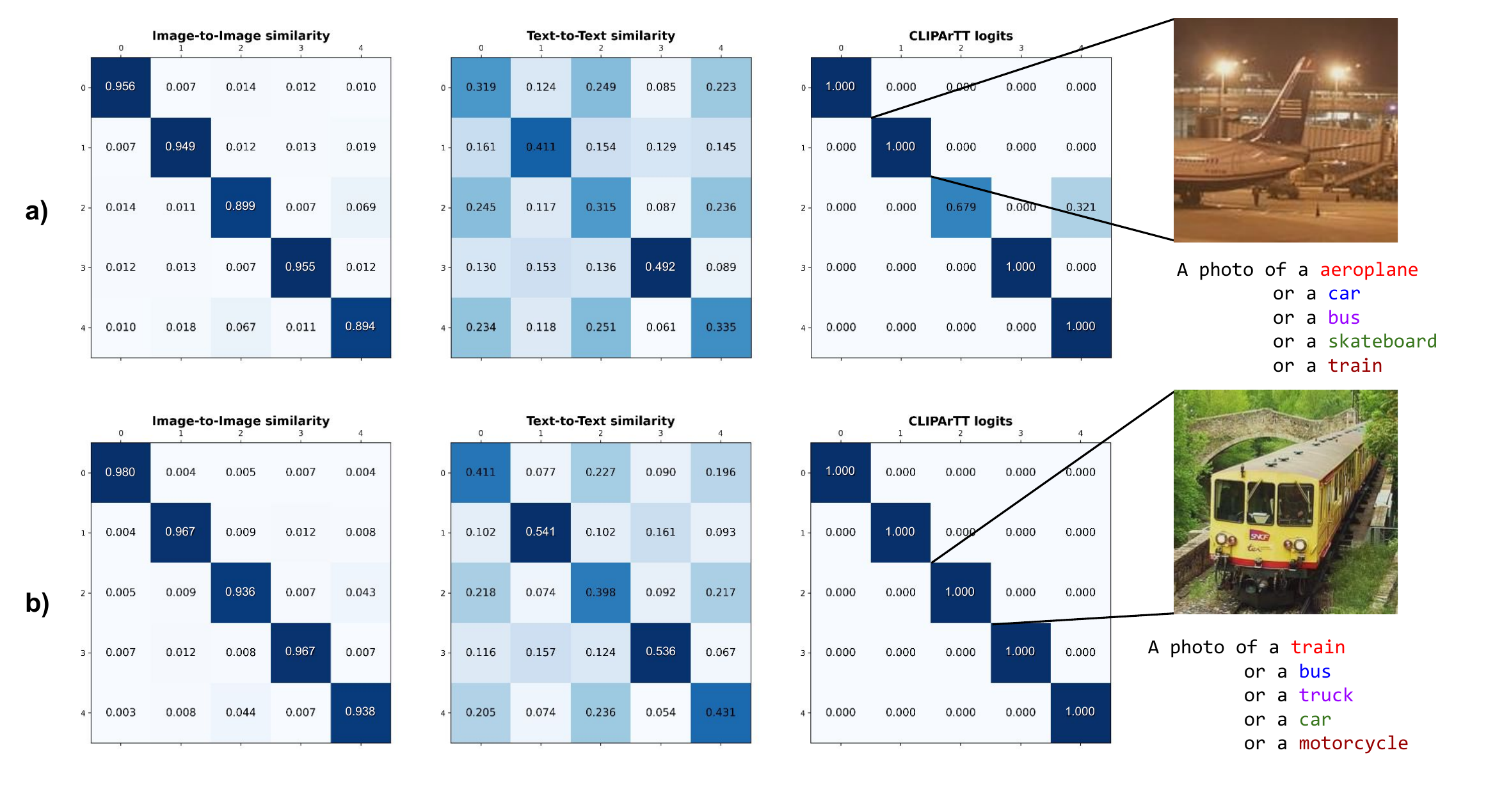}\\
    \end{tabular}
    \caption{\textbf{Top}: Example of similarity matrices ($\Sim^v$, $\Sim^t$) and CLIPArTT softmax probabilities ($\QQ$) for a batch of 5 examples and using $K=5$ classes.     
    \textbf{Bottom}: a) When using the identity matrix as pseudo-label for contrastive learning, the correct prediction is ambiguous, as the images are forced to both approaching and moving away from the right class. b) CLIPArTT uses soft pseudo-labels that smoothly guides the prediction towards the correct class by reducing the impact of ambiguities in the prompts.}
    \label{fig:similarity}    
    \vspace*{-10pt}
\end{figure*}

\subsection{CLIP-based classification}

Contrastive Language-Image Pre-training (CLIP) \cite{clip} consists of a visual encoder $f^v_{\theta}(\cdot)$, mapping an image $\xx$ to visual features $\zz^v \in \R^D$, and a text encoder $f^t_{\theta}(\cdot)$ transforming text prompts $\ttt$ to text features $\zz^t \in \R^D$. The visual and text encoders are trained jointly with a contrastive loss so that the feature embeddings of training images and their corresponding text prompt are close to each other, while those of different training examples are pushed apart. 

In a classification task with $K$ fixed classes, CLIP can be used to perform inference by encoding a pre-defined text prompt for each class, for example $\ttt_k =$ ``\texttt{a photo of a \{class $k$\}}''. For a new image $\xx_{i}$, the probability of belonging to class $k$ is then estimated based on cosine similarity,
\begin{equation}\label{eq:clip-classif}
    p_{ik} \, = \, \frac{\exp\big(\!\cos(\zz^v_{i}, \zz^t_k)/\tau\big)}{\sum_j \exp\big(\!\cos(\zz^v_{i}, \zz^t_j)/\tau\big)}, \ \
    \cos(\zz, \zz') = \frac{\tr{\zz}{\zz'}}{\|\zz\|_2 \!\cdot \! \|\zz'\|_2},
\end{equation}
where $\tau$ is a suitable softmax temperature.

The model in Eq. (\ref{eq:clip-classif}) can be used for test-time adaptation in various ways, the simplest one being entropy minimization as in TENT. This approach, which relies on the principle that the decision boundary lies in a low-density region of space giving rise to a low prediction entropy, adapts the model parameters by minimizing entropy on a test batch of size $B$:
\begin{equation}
\loss_{\mr{T\textsc{ent}}}(\theta) = -\frac{1}{B}\sum_{i=1}^B \sum_{k=1}^K p_{ik} \log p_{ik}
\end{equation}

However, this approach as well as similar ones based on pseudo-labels \cite{shot,cotta,pad} suffer from two important limitations. First, due to domain shifts, the model's prediction may be unreliable (e.g., giving the highest probability to the wrong class) and techniques such as entropy minimization or standard pseudo-label will only reinforce these errors during adaptation. Secondly, they assume that samples in a test batch are independent and do not directly leverage their semantic relationships.

\begin{table}[!ht]
    \centering    
    %\dorowcolors   
    \begin{small}
    \setlength{\tabcolsep}{5pt}
    \begin{tabular}{l|cc|cc}
    \toprule
        % ~ & CIFAR10 & ~ & CIFAR100 & ~ \\ \midrule
        ~ & \multicolumn{2}{c|}{CIFAR10} & \multicolumn{2}{c}{CIFAR100} \\ \cmidrule(l{4pt}r{4pt}){2-3} \cmidrule(lr){4-5}
        ~ & Top-1 & Top-3 & Top-1 & Top-3 \\ \midrule
        Original & 88.74 & 97.79 & 61.68 & 80.92 \\  \midrule
        Corrupted (-C) & 59.22 & 82.43 & 29.43 & 46.61 \\ \bottomrule
    \end{tabular}
    \end{small}
    \caption{Accuracy (\%) on CIFAR-10/100 and CIFAR-10/100-C datasets with Level 5 corruption for the top-1 or the top-3 predicted classes.}
	\label{tab:MotivationPrediction}
 %\vspace*{-10pt}
 \end{table}

\begin{figure}
        \centering
        \setlength{\tabcolsep}{3mm}
        \begin{tabular}{cc}
        \includegraphics[height=0.4\linewidth]{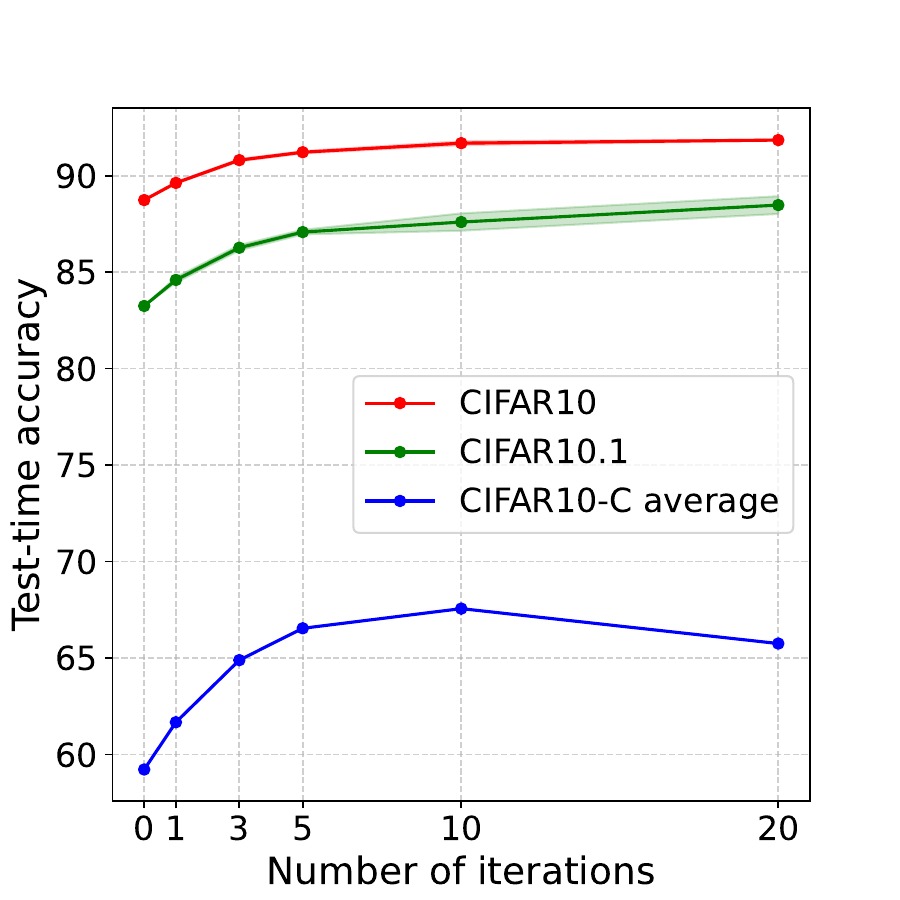} &          \includegraphics[height=0.4\linewidth]{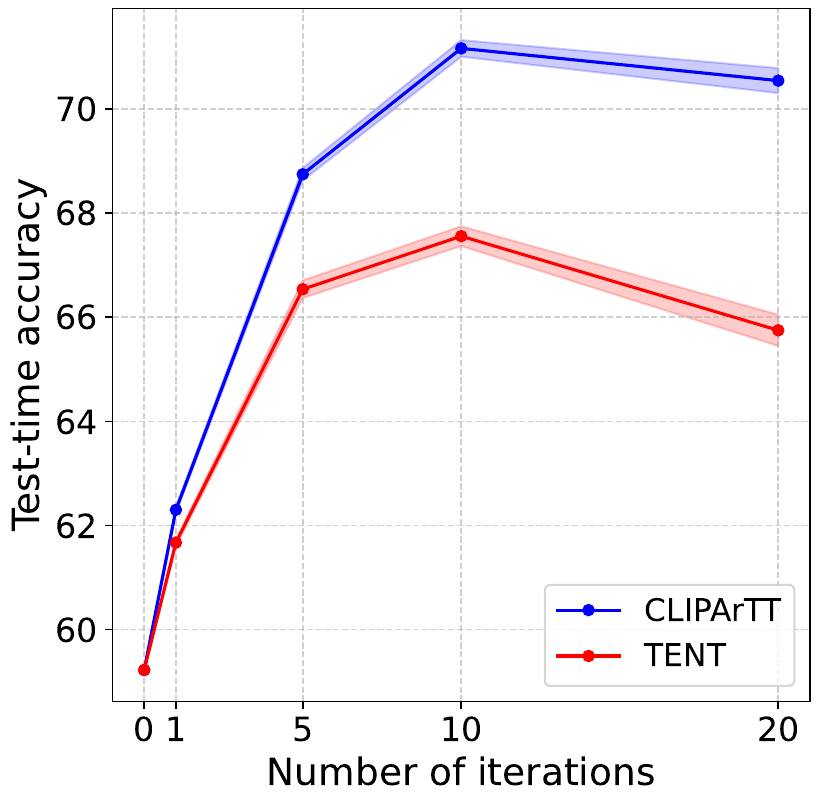}
        \end{tabular}
        \caption{Evolution of CLIPArTT's accuracy during test-time adaptation.
        \textbf{Left}: For different versions of CIFAR10. \textbf{Right}: Compared to TENT on CIFAR10-C.}
        \label{fig:iterations}    
        \vspace*{-10pt}
\end{figure}

\begin{comment}
\begin{figure}
    \begin{minipage}[b]{.2\textwidth}
        \centering
        \includegraphics[width=\linewidth]{Figures/TENTiter.pdf}
        \caption{Evolution of accuracy after several iterations of adaptation at test-time for TENT.}
        \label{fig:tent_iter} 
    \end{minipage}%
    \begin{minipage}[b]{.22\textwidth}
        \centering
        \includegraphics[width=\linewidth]{Figures/iter.pdf}
        \caption{CLIPArTT achieves higher accuracy in fewer iterations than TENT.}
        \label{fig:iterations}
    \end{minipage}
\end{figure}
\end{comment}

\subsection{Our CLIPArTT method}

The proposed CLIPArTT method (see Figure~\ref{fig:diagram}), addresses the above-mentioned limitations using two key insights. The first insight, inspired by conformal learning, is that the correct class is often among the top most probable ones, although the model's most confident prediction may not be always correct. This claim is supported by the results in Table~\ref{tab:MotivationPrediction} (first row: \emph{Original}), showing that the correct class is within the top-3 predictions 97.79\% of the times for CIFAR-10 (versus 88.74\% within the top-1), and 80.92\% of the times for CIFAR-100 (versus 61.68\% within the top-1). A way to include information about multiple classes could, therefore, help adapting the model at test time. The second insight is that the similarity between the batch samples could be evaluated based on their visual and/or text embeddings. As we will show below, such similarities could be exploited in a strategy related to Stochastic Neighbor Embedding (SNE) and graph-Laplacian regularization. 

\mypar{Instance-specific multi-class prompt.} Inspired by our first insight and recent work investigating CLIP’s ability to perform compositional logical reasoning \cite{brody2023potential}, we devise a novel technique to generate instance-specific prompts from the top-$k$ predictions, with $1 \leq k \ll K$. Specifically, for an image $\xx_i$, we estimate the CLIP-based class probabilities using Eq. (\ref{eq:clip-classif}) and then generate a new text prompt as $\hat{\ttt}_i =\,$\texttt{a photo of a \{class $i_{1}$\} or $\ldots$  or \{class $i_{k}$\}}'', where \texttt{\{class $i_{j}$\}} is the name of the class with $j$-th highest probability.

\mypar{Transductive TTA.} Next, we design a test-time adaptation loss that accounts for semantic relationships between batch samples. Let $\ZZ^v \in \real^{B \times D}$ and $\hat{\ZZ}^t \in \real^{B \times D}$ denote the \emph{normalized} visual and instance-specific text embeddings of the samples within the test batch, respectively. We compute an image-to-image similarity matrix, $\Sim^v = \ZZ^{v} \tr{(\ZZ^{v})} \in [-1,1]^{B \times B}$, and a text-to-text similarity matrix, $\Sim^t = \hat{\ZZ}^{t} \tr{(\hat{\ZZ}^{t})} \in [-1,1]^{B \times B}$. The former measures the affinity between each pair of samples within the batch in terms of their visual characteristics (shapes, textures, etc.). The latter captures common (or related) classes in the top-$k$ predictions of two samples, since it is computed using the instance-specific multi-class prompts. As illustrated in Figure \ref{fig:similarity} (\emph{top}), a broad range of similarity values are obtained with this approach.  

We deploy these two pairwise similarity matrices to compute pseudo-labels as follows:
\begin{equation}
    \label{Q-matrix}
    \QQ = \softmax\big((\Sim^v + \Sim^t)/2\tau\big) \in [0,1]^{B\times B} 
\end{equation}

where the softmax operation is applied column-wise and the temperature $\tau = 0.01$ in all our experiments.

Let $\hat{\PP}$ denote the zero-shot prediction matrix using our instance-specific multi-class text prompts:
\begin{equation}
\label{P-matrix}
\hat{\PP} = \softmax\big(\ZZ^v \tr{(\hat{\ZZ}^t)}/\tau\big),\,\ \hat{p}_{ij} = \frac{\exp\big(\!\cos(\zz^v_{i}, \hat{\zz}^t_j)/\tau\big)}{\sum_k \exp\big(\!\cos(\zz^v_{i}, \hat{\zz}^t_k)/\tau\big)}
\end{equation}

This matrix, along with the pairwise pseudo-labels we introduced in Eq. \eqref{Q-matrix}, yield our final TTA loss based on the cross-entropy:
\begin{equation}\label{eq:tta-loss}
\loss_{\mr{TTA}}(\theta) = -\frac{1}{B}\sum_{i=1}^B\sum_{j=1}^B q_{ij} \log \hat{p}_{ij}.
\end{equation}

Unlike recent approaches like TPT \cite{tpt}, which adapt CLIP by learning a text prompt, we instead update the normalization-layer parameters of the \emph{visual} encoder, which yields a light-weight, computationally efficient TTA method. We note that TPT requires to generate multiple augmentations during the forward pass, incurring a substantial overhead.  

The mechanism of CLIPArTT is illustrated in Fig.~\ref{fig:similarity}. Our soft pseudo-label reduces the risk of increasing the ambiguity of the predictions. Using the identity matrix (i.e., hard pseudo-labels) leads to uncertain class assignments, as the ambiguity of the text prompts would both attract and repel the right class if it is present in the same text prompt for two different images.

%After adaptation, the confidence of the image-to-text logits increases (i.e., the similarity matrix approaching the identity), with respect to before adaptation. This also results into a more accurate regrouping of images into their corresponding class cluster, which is most of the times explicitly encoded in the new text prompt. The image-to-image similarities guide the clustering of the images based on visual cues, thus regularizing the approximation to the right class.

\subsection{Link to existing techniques}

An important element of our TTA method is that it exploits the similarities between pairs of samples within the batch, in terms of the visual features and text prompts representing the top-$k$ classes. In this section, we draw a connection between the proposed method and two well-known techniques in machine learning: Stochastic Neighbor Embedding \cite{hinton2002stochastic} and graph-Laplacian regularization \cite{semi,label}.

\mypar{Stochastic Neighbor Embedding (SNE).} This popular technique for dimensionality reduction estimates the local probability of a point $\xx_j$ given a point $\xx_i$ based on their Euclidean distance $d_{ij} = \|\xx_i - \xx_j\|_2$:
\begin{equation}\label{eq:SNE}
p_{ij} \, = \, \frac{\exp\big(-\!d^2_{ij}\big)}{\sum_{k \neq i}\exp\big(-\!d^2_{ik}\big)} 
\end{equation}
The goal is to find, for each $\xx_i$, a low-dimensional embedding $\yy_i$ such that the local probabilities $q_{ij}$ computed using Eq. (\ref{eq:SNE}) on these embeddings, is similar to $p_{ij}$. This is achieved by minimizing the row-wise KL divergence between distribution matrices $\PP$ and $\QQ$. Our CLIPArTT method can be linked to SNE since $\cos(\xx_i, \xx_j) = -\!d^2_{ij} + 2$ when $\xx_i$, $\xx_j$ are unit normalized, and KL divergence between $\PP$ and $\QQ$ is equal to the cross-entropy between these matrices minus the entropy of $\PP$. In summary, our TTA loss in Eq. (\ref{eq:tta-loss}) ensures that the inter-modality (text-to-image) similarities of batch samples are aligned with their intra-modality ones (text-to-text and image-to-image).

\mypar{Graph-Laplacian regularization} is commonly-used in the context semi-supervised learning techniques \cite{semi} and label propagation \cite{label}. In our case, the Laplacian regularizer is computed over a set of $B$ nodes with predictions $\ZZ \in \real^{B \times D}$ as:
\begin{equation}
\loss_{\mr{reg}}(\ZZ) = \mr{trace}\big(\tr{\ZZ}\LL_{\mr{W}} \ZZ\big) \, = \, \sum_{i,j} w_{ij} \| \zz_i - \zz_j\|_2^2, 
\end{equation}
where $\LL_{\mr{W}} \in \real^{B \times B}$ 
is the Laplacian matrix defined from edge weight matrix $\WW$ as follows:\begin{equation}
[\LL_{\mr{W}}]_{ij} = \left\{\begin{array}{ll}
d_{ii} = \sum_j w_{ij}, & \text{ if } i=j \\
-w_{ij} & \text{ else}.
\end{array}
\right.
\end{equation}

This connection is made in the following proposition.

\begin{proposition}
The TTA loss in Eq. (\ref{eq:tta-loss}) can be expressed as a Laplacian regularization over a bipartite graph with one set of nodes for image embeddings and another for text embeddings. 
\begin{proof}
Expanding the $\softmax$ in Eq. (\ref{eq:tta-loss}), we get
\begin{align}
    \loss_{\mr{TTA}} & \, = \, 
    -\frac{1}{B}\sum_{i=1}^{B}\sum_{j=1}^{B}q_{ij} \log \frac{\exp\big(\tr{(\zz^{v}_i)} \zz^{t}_j/\tau\big)}{\sum_{k} \exp\big(\tr{(\zz^{v}_i)} \zz^{t}_k/\tau\big)} \\
    & \, = \, 
    \frac{1}{\tau B}\sum_{i}\sum_{j}q_{ij} \Big(-\!\tr{(\zz^{v}_i)} \zz^{t}_j \nonumber\\[-8pt]
    & \qquad\qquad\quad \ + \, \tau\underbrace{\log \sum_{k} \exp\big(\tr{(\zz^{v}_i)} \zz^{t}_k/\tau\big)}_{\small \text{LogSumExp (LSE)}}\Big)
\end{align} 

Since the feature embeddings are normalized, we have $\|\zz_i^v\|_2=\|\zz_j^t\|_2=1$ and thus $\tr{(\zz^{v}_i)} \zz^{t}_j = 2 - \|\zz^{v}_i- \zz^{t}_j\|_2^2$.
Moreover, following the scaled LogSumExp (LSE) rule, we can bound the right-side term as follows:
\begin{equation}
\max_{k}\,\{\tr{(\zz^{v}_i)} \zz^{t}_k\} \, < \, \tau LSE \, \leq \, \max_{k}\,\{\tr{(\zz^{v}_i)} \zz^{t}_k\} \, + \, \tau\log\!B
\end{equation}
For a small $\tau$ (we use a value of 0.01 in our method), the bounds tighten and we get that
\begin{equation}
LSE \, \approx \, \tfrac{1}{\tau}\max_{k}\,\{\tr{(\zz^{v}_i)} \zz^{t}_k\} \, = \, \tfrac{1}{\tau}\tr{(\zz^{v}_i)} \zz^{t}_i %\, \triangleq \, m^v_i
\end{equation} 

The last equality comes from the hypothesis that, in a well trained model, the maximum similarity occurs between the image embedding of a sample and its corresponding text embedding. Our TTA loss can then be expressed as
\begin{align}
    \loss_{\mr{TTA}} & \, \approx \,
    \frac{1}{\tau B}\Big(\sum_{i,j}  q_{ij} \|\zz^{v}_i- \zz^{t}_j\|_2^2 \, + \, \sum_i \tr{(\zz^{v}_i)} \zz^{t}_i \Big)  \, +    
    \text{\emph{const}} \nonumber\\
    & \, = \, \frac{1}{\tau B}\mr{trace}\big(\tr{(\ZZ^v)}\!(\LL_{\mr{Q}} + \II)\,\ZZ^t\big) \ +  \text{\emph{const}}
\end{align}
where $\II$ is the identity matrix. The modified Laplacian $\LL_{\mr{Q}} + \II$ enforces nodes with a high connection weight ($q_{ij}$) to have similar embeddings, while also avoiding embeddings to collapse into a single vector.
\end{proof}
\end{proposition}

\section{Experimental Settings}
\label{sec:experiments}

We evaluate CLIPArTT's performance across a variety of TTA datasets, covering scenarios such as natural images, common corruptions, simulated images, video, and Domain Generalization benchmark. This comprehensive framework allows for a robust assessment of the model's adaptability to various challenges, including domain shifts and corruptions. For a detailed description of the datasets, please refer to the supplementary materials.

\mypar{Test-time adaptation.} For test-time adaptation, model updates are applied to all the Layer Normalization (LN) layers within the visual encoder. We employ the ADAM optimizer with a fixed learning rate set to $10^{-3}$. Throughout our experiments, a consistent batch size of 128 is utilized to maintain uniformity and enable effective comparisons across different scenarios. As in previous TTA works, a smaller learning rate of $10^{-4}$ is preferred to adapt to the 3D renderings split \cite{clust3}, as it represents a more aggressive shift. 

\mypar{Benchmarking.} We compare CLIPArTT with \emph{state-of-the-art} methods. Specifically, we utilize adapted versions of TENT~\cite{tent} and LAME~\cite{lame} tailored to CLIP. While the classifier logits are used in previous TTA research involving CNNs, we follow the standard practice and use the image-to-text similarity to obtain the final logits in CLIP. Only the visual encoder is optimized where needed. TENT now updates only the affine parameters in LN layers through entropy minimization directly on these logits.
Notably, we employ 10 iterations for TENT adaptation, a choice informed by our findings, as evidenced in Fig~\ref{fig:iterations}. In LAME, the Laplacian regularizer is applied on the similarity between image features (obtained from the visual encoder). Finally, we include CLIP-tailored methods TTDN~\cite{TTDN}, TPT~\cite{tpt}, TDA~\cite{tda} and CALIP~\cite{calip}. Both TDA and CALIP require hyperparameter tuning for optimal performance. To ensure fairness in our benchmarking process, we opted to fix these hyperparameters as recommended for the ImageNet dataset, as suggested in their respective works. The batch-size for TPT is reduced to 32 due to its reliance on image augmentations and high memory requirements.

\begin{table}[!t]
    \centering    
    %\dorowcolors   

    \begin{small}
    \setlength{\tabcolsep}{5pt}
    \begin{tabular}{l|ccc}
    \toprule
        & $K$\,=\,1 & $K$\,=\,3 & $K$\,=\,4 \\ \midrule
        CIFAR-10 & 89.80\ppm0.05 & 90.04\ppm0.13 & \textbf{90.41\ppm0.07} \\ %\midrule
        CIFAR-10.1 & 85.37\ppm0.17 & \textbf{86.35\ppm0.27} & 86.07\ppm0.21 \\ 
        %\midrule
        CIFAR-10-C & 70.79\ppm0.15 & \textbf{71.17\ppm0.16} & 70.99\ppm0.15 \\ \bottomrule
    \end{tabular}
    \end{small}
    %\vspace*{-1pt}
    \caption{Accuracy (\%) on CIFAR-10, CIFAR-10.1 and CIFAR-10-C datasets with Level 5 corruption for different number of $K$ selected classes to create pseudo-labels.}
	\label{tab:KCifar10}
    %\vspace*{-3pt}
\end{table}

\begin{table}[!t]
    \centering
    \resizebox{\linewidth}{!}{
    \setlength{\tabcolsep}{4pt}
    \begin{small}
    \begin{tabular}{l|cccc}
    \toprule
        & Iter\,=\,1 & Iter\,=\,5 & Iter\,=\,10 & Iter = 20 \\ \midrule
        CIFAR-10 & 89.59\ppm0.01 & \textbf{90.54\ppm0.09} & 90.04\ppm0.13 & 88.32\ppm0.12 \\ 
        %\midrule
        CIFAR-10.1 & 84.78\ppm0.02 & \textbf{86.67\ppm0.06} & 86.35\ppm0.27 & 84.33\ppm0.31 \\ 
        %\midrule
        CIFAR-10-C & 62.30\ppm0.06 & 68.75\ppm0.12 & \textbf{71.17\ppm0.16} & 70.55\ppm0.24 \\ \bottomrule
    \end{tabular}
    \end{small}
    }
    %\vspace*{-3pt}
    \caption{Accuracy (\%) on CIFAR-10, CIFAR-10.1 and CIFAR-10-C datasets with Level 5 corruption for different number of iterations to update the model at test-time.}
	\label{tab:iterCifar10}
    %\vspace*{-10pt}
\end{table}

\section{Results}
\label{sec:results}

In this section, we outline the experimental outcomes derived from CLIPArTT and provide a comprehensive analysis. We first show a series of exploratory ablations that later help conduct the final experiments on the different datasets and compare with \emph{state-of-the-art}.

\subsection{Ablation Studies}

\vspace*{-3pt}
\mypar{Number of classes for new prompts.} Determining the number of classes in building new prompts presents a critical aspect in our methodology. Table~\ref{tab:MotivationPrediction} illustrates various scenarios where $K\!=\!3$ emerges as a favorable choice, exhibiting a remarkable accuracy above $90\%$ across multiple instances of CIFAR-10-C datasets. This initial observation is further corroborated by the findings presented in Table~\ref{tab:KCifar10}. However, the decision becomes more nuanced when applied to CIFAR-100 datasets. While leveraging the top 3 classes contributes to enhanced accuracy, it does not guarantee optimal performance. Similarly, adopting a strategy akin to CIFAR-10, wherein $30\%$ of the classes are chosen, proves impractical due to resulting lengthy sentences that are impossible to tokenize. Nonetheless, our analysis shows that $K\!=\!3$ consistently yields superior performance, particularly evident in the average results across CIFAR-100-C. Consequently, we maintain $K\!=\!3$ as the preferred configuration for all subsequent experiments on CIFAR datasets.

\begin{table}[!t]
    \centering
    % \rowcolors{4}{white}{gray!15}
    \resizebox{.9\linewidth}{!}{
    \setlength{\tabcolsep}{5pt}
    \begin{tabular}{l|c|ccc}
    \toprule
        & Avg. Prompt & Image & Text & Image\,+\,Text \\ 
        \midrule
        CIFAR-10 & 76.94\ppm0.41 & \textbf{90.18\ppm0.02} & 89.05\ppm0.14 & 90.04\ppm0.13 \\ 
        %\midrule
        CIFAR 10.1 & 67.83\ppm0.49 & 86.25\ppm0.37 & 84.85\ppm0.40 & \textbf{86.35\ppm0.27} \\ 
        CIFAR-10-C & 43.08\ppm0.08 & 70.98\ppm0.15 & 70.03\ppm0.21 & \textbf{71.17\ppm0.16} \\ \bottomrule
    \end{tabular}
    }
    %\vspace*{-3pt}
    \caption{Accuracy (\%) on CIFAR-10, CIFAR-10.1 and CIFAR-10-C datasets with different targets.}
	\label{tab:targetCifar10}
%\vspace*{-3pt}
\end{table}

\begin{table}[!t]
    \centering
    % \begin{footnotesize}
    %\dorowcolors
    \resizebox{\linewidth}{!}{
    %\begin{small}
    \setlength{\tabcolsep}{4pt}
    \begin{tabular}{l|c|cccc}
    \toprule
        & CLIP & %BS\,=\,8 & 
        BS\,=\,16 & BS\,=\,32 & BS\,=\,64 & BS\,=\,128 \\ \midrule
        CIFAR-10 & 88.74 & %82.50\ppm0.13 & 
        85.89\ppm0.19 & 88.25\ppm0.15 & 89.48\ppm0.15 & 90.04\ppm0.13 \\ 
        %\midrule
        CIFAR-10.1 & 
        83.25 & 
        %77.20\ppm0.92 & 
        81.55\ppm0.53 & 84.00\ppm0.31 & 85.40\ppm0.08 & 86.35\ppm0.27 \\ 
        %\midrule
        CIFAR-10-C & 
        59.22 & 
        %61.10 & 
        64.72\ppm0.23 & 67.70\ppm0.23 & 69.82\ppm0.20 & 71.17\ppm0.16 \\ \bottomrule
    \end{tabular}
    %\end{small}
    }
    % \end{footnotesize}
    %\vspace*{-3pt}
    \caption{Accuracy (\%) on CIFAR-10, CIFAR-10.1 and CIFAR-10-C datasets with Level 5 corruption for different number of batch-size.}
	\label{tab:BSCIfar10}
 %\vspace*{-10pt}
\end{table}

\begin{comment}
\begin{table}[!h]
    \centering
    \resizebox{\linewidth}{!}{
    \setlength{\tabcolsep}{3pt}
    \begin{tabular}{l|cccccc}
    \toprule
        ~ & $K$\,=\,1 & $K$\,=\,3 & $K$\,=\,5 & $K$\,=\,7 & $K$\,=\,10 & $K$\,=\,20 \\ 
        \midrule
        CIFAR-100 & 69.00\ppm0.22 & 69.79\ppm0.04 & 69.68\ppm0.07 & 69.56\ppm0.02 & 69.78\ppm0.02 & \textbf{69.93\ppm0.08} \\ \midrule
        CIFAR-100-C & 40.75 & \textbf{41.51} & 41.21 & 41.09 & 41.16 & 40.89 \\ \bottomrule
    \end{tabular}
    }
    \vspace{5pt}
    \caption{Accuracy (\%) on CIFAR-100 and CIFAR-100-C datasets with Level 5 corruption for different number of $K$ selected classes to create pseudo-labels.}
	\label{tab:KCIfar100}
    %\vspace*{-15pt}
\end{table}
\end{comment}

\begin{comment}
\begin{table}[!t]
    \centering
    \resizebox{\linewidth}{!}{
    \setlength{\tabcolsep}{3pt}
    \begin{tabular}{l|ccccc}
    \toprule
        ~ & $K$\,=\,1 & $K$\,=\,3 & $K$\,=\,5 %& $K$\,=\,7 
        & $K$\,=\,10 & $K$\,=\,20 \\ 
        \midrule
        CIFAR-100 & 69.00\ppm0.22 & 69.79\ppm0.04 & 69.68\ppm0.07 
        %& 69.56\ppm0.02 
        & 69.78\ppm0.02 & \textbf{69.93\ppm0.08} \\ 
        %\midrule
        CIFAR-100-C & 40.75 & \textbf{41.51} & 41.21 
        & 41.09 
        %& 41.16 
        & 40.89 \\ \bottomrule
    \end{tabular}
    }
    \caption{Accuracy (\%) on CIFAR-100 and CIFAR-100-C datasets with Level 5 corruption for different number of $K$ selected classes to create pseudo-labels.}
	\label{tab:KCIfar100}
    %%\vspace*{-15pt}
\end{table}
\end{comment}

\begin{table*}[!h]
    \centering
    %\rowcolors{4}{white}{gray!15}
    \begin{small}
    \setlength{\tabcolsep}{5pt}
    \begin{tabular}{l|cccc|cccc}
    \toprule
       & \multicolumn{4}{c|}{CIFAR-100-C} & \multicolumn{4}{c}{ImageNet-C} \\
        \cmidrule(l{3pt}r{3pt}){2-5}       \cmidrule(l{3pt}r{3pt}){6-9}
        & CLIP & TENT & TDA & CLIPArTT & CLIP & TENT & TDA & CLIPArTT \\ \midrule
        Gaussian Noise & 14.80 & 14.38\ppm0.14 & \phz{}8.20\ppm0.35 & \textbf{25.32\ppm0.14} & 12.27 & 12.28\ppm0.05 & 11.54\ppm0.05 & \textbf{20.18\ppm0.61} \\ 
        Shot noise & 16.03 & 17.34\ppm0.27 & \phz{}9.58\ppm0.43 & \textbf{27.90\ppm0.05} & 12.20 & \phz{}8.46\ppm0.08 & 12.13\ppm0.13 & \textbf{20.94\ppm0.08} \\ 
        Impulse Noise & 13.85 & 10.03\ppm0.13 & \phz{}7.63\ppm0.19 & \textbf{25.62\ppm0.09} & 12.89 & 15.71\ppm0.04 & 12.12\ppm0.08 & \textbf{19.95\ppm0.02} \\ 
        Defocus blur & 36.74 & 49.05\ppm0.07 & 25.59\ppm0.41 & \textbf{49.88\ppm0.23} & 21.60 & 20.61\ppm7.05 & 21.39\ppm0.08 & \textbf{24.51\ppm0.13} \\ 
        Glass blur & 14.19 & \phz{}3.71\ppm0.07 & \phz{}9.83\ppm0.56 & \textbf{27.89\ppm0.03} & 10.84 & 17.20\ppm0.08 & 10.74\ppm0.06 & \textbf{19.51\ppm0.02} \\ 
        Motion blur & 36.14 & 46.62\ppm0.27 & 28.92\ppm0.18 & \textbf{47.93\ppm0.14} & 17.85 & 24.60\ppm0.04 & 18.45\ppm0.06 & \textbf{25.80\ppm1.16} \\ 
        Zoom blur & 40.24 & 51.84\ppm0.15 & 31.08\ppm0.36 & \textbf{52.70\ppm0.06} & 16.38 & 21.13\ppm0.08 & 16.83\ppm0.03 & \textbf{24.00\ppm0.06} \\ 
        Snow & 38.95 & 46.71\ppm0.21 & 32.94\ppm0.12 & \textbf{49.72\ppm0.01} & 21.91 & 24.19\ppm0.09 & 22.97\ppm0.05 & \textbf{27.26\ppm0.06} \\ 
        Frost & 40.56 & 44.90\ppm0.27 & 34.84\ppm0.25 & \textbf{49.63\ppm0.12} & 23.33 & 23.34\ppm0.06 & 24.68\ppm0.06 & \textbf{26.56\ppm0.08} \\ 
        Fog & 38.00 & 47.31\ppm0.04 & 31.13\ppm0.15 & \textbf{48.77\ppm0.04} & 26.39 & 28.58\ppm0.04 & 27.72\ppm0.05 & \textbf{34.68\ppm0.02} \\ 
        Brightness & 48.18 & 60.58\ppm0.18 & 42.36\ppm0.10 & \textbf{61.27\ppm0.08} & 45.87 & 47.16\ppm0.01 & 48.06\ppm0.04 & \textbf{47.21\ppm0.07} \\ 
        Contrast & 29.53 & 45.90\ppm0.11 & 18.03\ppm0.07 & \textbf{48.55\ppm0.24} & 16.16 & 21.62\ppm0.01 & 16.09\ppm0.05 & \textbf{23.38\ppm0.12} \\ 
        Elastic transform & 26.33 & 33.09\ppm0.08 & 18.88\ppm0.24 & \textbf{37.45\ppm0.08} & 16.04 & 17.17\ppm0.01 & 17.75\ppm0.01 & \textbf{23.95\ppm0.04} \\ 
        Pixelate & 21.98 & 26.47\ppm0.09 & 14.59\ppm0.30 & \textbf{33.88\ppm0.14} & 28.00 & 33.56\ppm1.71 & 30.03\ppm0.05 & \textbf{34.12\ppm0.09} \\ 
        JPEG compression & 25.91 & 29.89\ppm0.07 & 17.56\ppm0.11 & \textbf{36.07\ppm0.32} & 27.44 & 31.89\ppm0.02 & 29.36\ppm0.01 & \textbf{32.23\ppm0.12} \\ \midrule
        Average & 29.43 & 35.19 & 22.08 & \textbf{41.51} & 20.61 & 23.17 & 21.32 & \textbf{26.95} \\ \bottomrule
    \end{tabular}
    \end{small}
    %\vspace*{-3pt}
    \caption{Accuracy (\%) on CIFAR-100-C and ImageNet-C datasets with ViT-B/32 as visual encoder.}
    \label{tab:B32Cifar100andImagenet}
    %\vspace*{-15pt}
\end{table*}

\mypar{Comparison against prompt averaging.} The combination of $K$ classes inside the template  \{class $i_{1}$\} or $\ldots$  or \{class $i_{k}$\}, poses a sensible question: how comparable is this alternative against a simple prompt averaging? In CLIPArTT, we hypothesize that combining the most probable classes' names in natural language, helps to exploit CLIP's langauge understanding more effectively. Experiments conducted on CIFAR-10 and its variants are shown in Table~\ref{tab:targetCifar10}, showing that our template, while simple, gives rise to an important accuracy gain with respect to prompt averaging.

\mypar{Number of iterations.} In line with our approach for TENT, we investigate optimal number of iterations. As demonstrated in Table~\ref{tab:iterCifar10}, iterating 10 times strikes the most favorable balance, exhibiting superior performance across the overall average on CIFAR-10-C and yielding optimal results for numerous corruption types. Nonetheless, it is imperative to acknowledge that employing a lower number of iterations (e.g., Iter\,=\,5) may yield improved outcomes in scenarios with minimal or no distribution shift, while a higher number of iterations (e.g., Iter\,=\,20) may be more effective in mitigating severe distribution shifts. As a result, 10 iterations are used for all forthcoming experiments.

\mypar{Pseudo-label selection for adaptation.} For the target, we employed a linear combination of both image-to-image and text-to-text similarities to encapsulate the information derived from both modalities.
This combination proved to be more effective than each modality separately (see Table~\ref{tab:targetCifar10}). While selecting solely the text-to-text similarity for adaptation performs well, it falls short of achieving optimal results. Conversely, utilizing only the image-to-image similarity mainly improves for the original CIFAR-10 dataset and few corruption types.

\mypar{Performance of the model for different batch-sizes.} TTA methods have historically exhibited limitations when applied with small batch sizes. In our study, we investigate the performance of our model in this regard. As depicted in Table~\ref{tab:BSCIfar10}, performance improves significantly with increasing batch size. However, beyond a batch size of 8, no further performance gains are observed. This phenomenon can be attributed to the fact that, with smaller batch sizes, our method can potentially introduce uncertainty by using multiple classes, thereby diminishing performance, especially when the model is already confident. In subsequent experiments, we maintain a batch size of 128, consistent with prevailing practices in the \emph{state-of-the-art} methodologies.

%\vspace{-5pt}
\subsection{Comparison on different datasets}
\vspace*{-2pt}
In standard TTA, the concept of domain shift is strictly related to the source dataset used for pretraining. In the context of CLIP, the presence of domain shifts is less evident, as the source data contained an enormous amount of images that likely include different domains. For this reason, we categorize our experiments per dataset type.

\begin{table*}[!h]
    \centering
    %\rowcolors{4}{white}{gray!15}
    \begin{small}
    \setlength{\tabcolsep}{5pt}
    % \begin{threeparttable}
    \begin{tabular}{l|c|cccccccc}
    \toprule
         & Shift & CLIP & LAME & TENT & TPT\,{\scriptsize(BS=32)} & TTDN & TDA & CALIP & CLIPArTT \\ \midrule
         
        CIFAR-10 & \xmark & 88.74 & 89.36\ppm0.06 & \textbf{91.69\ppm0.10} & 88.06\ppm0.06 & 89.06\ppm0.02 & 84.09\ppm0.04 & 86.79 \ppm 0.01 &90.04\ppm0.13 \\
        CIFAR-10.1 & \xmark & 83.25 & 81.22\ppm0.33 & \textbf{87.60\ppm0.45} & 81.80\ppm0.27 & 83.77\ppm0.06 & 78.98\ppm0.37 & 81.02 \ppm0.03 & 86.35\ppm0.27 \\   
        CIFAR-100 & \xmark & 61.68 & 58.27\ppm0.17 & 69.74\ppm0.16 & 63.78\ppm0.28 & 64.10\ppm0.05 & 60.32\ppm0.06 & 61.94\ppm0.01 &\textbf{69.79\ppm0.04} \\
        ImageNet & \xmark & 61.81 & 61.17\ppm0.02  & 62.13\ppm0.02 & 60.74\ppm0.16
         &61.85\ppm0.02  & \textbf{64.49\ppm0.01} & 58.50\ppm0.00 & 61.39\ppm 0.09 \\
         VisDA-C (YT) & \xmark & 84.45 & 84.72\ppm0.01  & %\textbf{85.46\ppm0.02} & 
        84.41\ppm0.02 & 82.95\ppm0.01 & 83.29\ppm0.03 & 84.47\ppm0.08 & \textbf{85.02\ppm 0.00} &83.46\ppm0.03\\
        PACS & \xmark & 93.65 & 93.68\ppm0.04  & 93.81\ppm0.03 & 93.23\ppm0.12 & \textbf{94.77\ppm0.07}  & 92.94\ppm1.08 & 94.11\ppm0.00 & 93.95\ppm 0.06 \\
        Office Home & \xmark & 77.53 & 75.46\ppm0.12  & 77.68\ppm0.06 & 77.20\ppm0.24 & \textbf{79.76\ppm0.04}  & 77.70\ppm0.13 & 78.75\ppm0.01 & 77.56\ppm 0.09 \\        
        \midrule
        CIFAR-10-C & \cmark & 59.22 & 50.57\ppm0.30 & 67.56\ppm0.19 & 56.80\ppm0.22 & 60.13\ppm0.07 & 48.00\ppm0.68 & 56.58\ppm0.0 &\textbf{71.17\ppm0.16} \\       
        CIFAR-100-C & \cmark & 29.43 & 26.23\ppm0.12 & 35.19\ppm0.16 & 30.46\ppm0.14 & 31.90\ppm0.04 & 22.08\ppm0.29 & 29.92\ppm0.01 & \textbf{41.51\ppm0.15} \\        
        ImageNet-C & \cmark & 20.61 & 20.02\ppm0.16 & 23.17\ppm0.54 & 21.10\ppm0.06 & 21.28\ppm0.02 & 21.32\ppm0.06 & 21.00\ppm0.00 &\textbf{26.95\ppm0.12} \\ 
        VisDA-C (3D) & \cmark & 84.51 & 84.75\ppm0.17 & 
        % 85.82\ppm0.02 &
        83.85\ppm0.03 & 78.45\ppm0.05 & 80.45\ppm0.01 & 84.22\ppm0.10 & 83.81\ppm0.08&\textbf{87.24\ppm0.04} \\
        \bottomrule
    \end{tabular}
    % \begin{tablenotes}
    %     \item[1] Too heavy to compute.
    % \end{tablenotes}
    % \end{threeparttable}
    \end{small}
    \vspace*{-3pt}
    \caption{Accuracy (\%) on CIFAR-10, CIFAR-10.1 and CIFAR-10-C datasets with ViT-B/32 as visual encoder. The Shift column denotes a domain shift from standard images recognized by CLIP: corruption in CIFAR-10-C, CIFAR-100-C and ImageNet-C datasets, synthetic images in VisDA-C (3D).}
	\label{tab:B32Cifar10}
% \vspace*{-15pt}
\end{table*}

\mypar{Natural images.} In the different Tables~\ref{tab:B32Cifar10} and \ref{tab:Cifar10}, CLIPArTT consistently enhances accuracy across CIFAR-10, CIFAR10.1, and CIFAR-100 datasets compared to the baseline (+2\%, +3\%, +8\% respectively with ViT-B/32). This improvement suggests that uncorrupted datasets, which may not have been seen by the model during training, can benefit from adaptation. Consequently, a zero-shot model can leverage adaptation at test-time, relying solely on the model's predictions. However, while CLIPArTT outperforms the baseline, as well as LAME, TPT and TTDN, it is worth noting that TENT may yield superior results depending on the visual encoder and the number of classes. Entropy minimization used solely on confident results proves to be also effective. Interestingly, the corruptions in which TENT achieves a better performance tend to already have higher accuracy. For a larger dataset such as ImageNet, most methods perform similarly, with the exception of TDA and CALIP, which improve CLIP performance by over 2\%. Notably, both TDA and CALIP require hyperparameter tuning, and we used the recommended settings optimized for ImageNet.

\mypar{Varied styles and textures.} Next we compared methods on two datasets, PACS and OfficeHome, which contain images from very different domains and are often used as benchmark in domain generalization. Once again, these domains do not constitute real distribution shifts for CLIP which was trained with a broad range of datasets. As can be seen, the best performance is achieved by methods such as TTDN for this setting. Nevertheless, it is important to highlight CLIPArTT's robustness, as it consistently maintains or improves model performance after adaptation. For instance, on the PACS dataset, CLIPArTT achieves an accuracy of 93.95\%, compared to 93.65\% for the original CLIP.

\begin{table}[!h]
\centering
\setlength{\tabcolsep}{5pt}
\begin{small}
\begin{tabular}{l|c|ccc}
\toprule
%\rowcolor{white}
& Backbone     & CLIP  & TENT  & %TPT\,{\scriptsize(BS=32)} & 
CLIPArTT\\ \midrule %\rowcolor{white}
%\multirow{2}{*}{CIFAR-10}   & \multicolumn{1}{c|}{ViT-B/16} & 89.25 & \textbf{92.75\ppm0.17} & 89.80\ppm0.05       & 92.61\ppm0.05          \\ 
% & \multicolumn{1}{c|}{ViT-L/14} & 95.36 & \textbf{96.13\ppm0.06} & 95.18\ppm0.02       & 95.16\ppm0.03          \\ \midrule  %\rowcolor{white}
\multirow{2}{*}{CIFAR-10.1} & \multicolumn{1}{c|}{ViT-B/16} & 84.00 & 88.52\ppm0.33          & %83.75\ppm0.21       & 
\textbf{88.72\ppm0.33} \\
 & \multicolumn{1}{c|}{ViT-L/14} & 91.20 & \textbf{92.22\ppm0.25} & %91.32\ppm0.12       & 
 91.02\ppm0.02          \\ \midrule %\rowcolor{white}
\multirow{2}{*}{CIFAR-10-C}                   & \multicolumn{1}{c|}{ViT-B/16} & 60.15 & 68.00\ppm0.18          & %59.75       & 
\textbf{73.22\ppm0.17} \\ 
& \multicolumn{1}{c|}{ViT-L/14} & 76.04 & \textbf{79.18\ppm0.10} & 
%75.01       & 
78.06\ppm0.15          \\ \midrule
%\multirow{2}{*}{CIFAR-100}  & \multicolumn{1}{c|}{ViT-B/16} & 64.76 & \textbf{71.73\ppm0.14} & 67.15\ppm0.23 & 71.34\ppm0.07 \\
%& \multicolumn{1}{c|}{ViT-L/14} & 73.28 & 78.03\ppm0.08 & 76.85\ppm0.06& \textbf{79.42\ppm0.08} \\ \midrule \rowcolor{white}
\multirow{2}{*}{CIFAR-100-C} & \multicolumn{1}{c|}{ViT-B/16} & 32.01 & 37.90\ppm0.18 & 
%33.73 & 
\textbf{40.08\ppm0.18} \\ 
  & \multicolumn{1}{c|}{ViT-L/14} & 44.59 & 50.14\ppm0.15 & %47.58 & 
  \textbf{52.52\ppm0.18} \\ \bottomrule
\end{tabular}
\end{small}
%\vspace{5pt}
%\vspace*{-3pt}
\caption{Accuracy (\%) on CIFAR-10, CIFAR-10.1, CIFAR-10-C, CIFAR-100 and CIFAR-100-C datasets with ViT-B/16 and ViT-L/14 as visual encoders.}
\label{tab:Cifar10}
%\vspace*{-12pt}
\end{table}

\begin{comment}
\begin{table*}[h!]
\centering
%\dorowcolors
\setlength{\tabcolsep}{4pt}
\begin{tabular}{l|l|cccccc}
\toprule
%\rowcolor{white}
& Backbone  & CLIP  & TENT &  TENT-10&     LAME      & TPT\,{\scriptsize(BS=32)} &    CLIPArTT   \\ \midrule  %\rowcolor{white}
\multirow{3}{*}{\textsc{3D}} & ViT-B/16 & 87.04 & 87.15\ppm0.21  & \textbf{88.08\ppm0.05} & 85.83\ppm0.06 & 83.85\ppm0.06 & 88.01\ppm0.47 \\ 
 & ViT-B/32 & 84.51 & 84.75\ppm0.17  & 85.82\ppm0.02 & 83.85\ppm0.03 & 78.45\ppm0.05 & \textbf{87.24\ppm0.04} \\ %\rowcolor{white}
 & ViT-L/14 & 90.62 & 90.93\ppm0.01  & \textbf{91.33\ppm0.01} & 90.15\ppm0.07 & 91.07\ppm0.05 & 91.18\ppm0.01 \\ \midrule %\rowcolor{white}
\multirow{3}{*}{\textsc{YT}} & ViT-B/16 & 86.71 & 86.96\ppm0.02  & \textbf{87.44\ppm0.02} & 86.24\ppm0.0  & 85.69\ppm0.05 & 86.22\ppm0.01 \\ 
 & ViT-B/32 & 84.45 & 84.72\ppm0.01  & \textbf{85.46\ppm0.02} & 84.41\ppm0.02 & 82.95\ppm0.01 & 83.46\ppm0.03 \\ \rowcolor{white}
 & ViT-L/14 & 84.93 & 85.38\ppm0.01  & \textbf{86.65\ppm0.05} & 85.08\ppm0.04 & 85.54\ppm0.38 & 85.55\ppm0.01 \\ \bottomrule
\end{tabular}
%\vspace{5pt}
\caption{Accuracy (\%) on the VisDA-C training split (3D) and validation split (YT) with the three different feature extractors.}
\label{tab:VisDA}
%%\vspace*{-15pt}
\end{table*}
\end{comment}

\mypar{Common corruptions.} In the various referenced tables, including Tables~\ref{tab:B32Cifar10} and \ref{tab:Cifar10}, CLIPArTT consistently outperforms the most benchmarks across all visual encoders, resulting in enhancements of up to 13\% on average. A more detailed examination reveals that CLIPArTT frequently surpasses TENT, particularly in instances where the baseline performs poorly. This observation implies that CLIPArTT consistently outperforms TENT, which tends to compromise the model's performance under conditions of low baseline confidence. For instance, in Table~\ref{tab:B32Cifar100andImagenet}, TENT experiences a 3\% decrease compared to the baseline under \emph{Impulse Noise}, while CLIPArTT exhibits a 12\% improvement. In Figure~\ref{fig:iterations} (\emph{right}), it is evident that CLIPArTT achieves better performance more rapidly across varying numbers of iterations. Conversely, when the baseline is confident, CLIPArTT consistently maintains a high performance. Finally, our method performs effectively on larger datasets with a greater number of classes, consistently improving upon the baseline by 6\% on ImageNet-C.

\mypar{Simulated images and video.} 
The performance of CLIPArTT on the YT dataset is suboptimal, a trend that can be observed in scenarios where the CLIP model already exhibits high confidence. However, our observations indicate that on the 3D dataset, CLIPArTT demonstrates a notable competitive advantage over other Test-Time Adaptation (TTA) methods and the baseline, achieving an improvement of nearly 3\%. Additionally, while TENT -- often considered one of the more robust methods in handling various corruptions -- results in a performance decline relative to the baseline, CLIPArTT maintains its effectiveness.

% Results on the 3D and YT splits of VisDA-C are shown in Table~\ref{tab:B32Cifar10}. With respect to pure CLIP, our method achieves an important improvement of accuracy with the three different image encoders. CLIPArTT also achieves a competitive advantage against TENT, LAME and TPT. We observed however that using TENT with 10 iterations (here named TENT-10) is a strong contender, achieving practically the same score on 3D images and outperforming on YouTube frames by a small margin. This can mean that TENT's entropy minimization can achieve faster improvement with certain domain shifts. 

\subsection{Limitations}
Our results indicate that CLIPArTT performs particularly well under significant domain shifts such as corruptions, though it is slightly less effective on natural images. While the performance of CLIPArTT in these cases is not as strong as that of TENT, it remains stable without deteriorating CLIP’s capabilities, whereas the  performance of TENT degrades under severe domain shifts. It is important to mention that any dataset not included in CLIP's pre-training can be considered a potential domain shift. However, natural images are more likely to have been seen by CLIP during pre-training, making adaptation for them prone to overfitting. This may explain why CLIPArTT excels in more challenging scenarios. To further explore the potential limitations of our method, we have included two additional scenarios in the supplementary material: \emph{a)} evaluation in a highly imbalanced setting, where only $C$ randomly chosen classes are present in a batch at random, and \emph{b)} open-set classification, where out-of-distribution images (i.e., from classes not present in the text prompts) are introduced into the batches.

\section{Conclusions}
\label{sec:conclusions}

We introduce CLIPArTT, a novel Test-Time Adaptation framework designed specifically for VLMs. By leveraging the model's predictions as new pseudo-labels, we effectively minimize cross-entropy loss at test-time, thereby enhancing model performance even within a zero-shot setting. To ensure fair comparisons, we also incorporate Test-Time Adaptation models into this new benchmark for VLMs.

A comprehensive ablation study determined optimal hyperparameters and helped gain deeper insights into the various model configurations. Our experimental results demonstrate that CLIPArTT achieves highly competitive performance across TTA datasets, surpassing \emph{state-of-the-art} approaches. While TENT remains a strong competitor, our model offers enhanced versatility by effectively addressing both natural and severe domain shifts, thus exhibiting robustness across various scenarios.

Exploring the potential of text prompts in classification is a promising avenue for future research. Investigating alternative methods to fine-tune these text prompts could yield valuable insights. Moreover, extending the study of Test-Time Adaptation to other scenarios, such as segmentation or object detection with Vision-Language Models (VLMs), holds significant promise for advancing our understanding of model adaptability and performance across diverse tasks.

\newpage

\section*{\centering CLIPArTT: Light-weight Adaptation of CLIP to New Domains at Test Time -- Supplementary Material}

\appendix

\section{Dataset Details}
We evaluate CLIPArTT's performance across diverse TTA datasets using established methodologies. These datasets simulate challenging scenarios, offering insights into our approach's efficacy. Additionally, we explore CLIPArTT's adaptability on other datasets through zero-shot test-time adaptation.

Our evaluation framework encompasses \emph{natural images}, \emph{varied styles and textures images}, \emph{common corruptions},  \emph{simulated images}, and \emph{video} providing a comprehensive assessment of the model's performance across diverse challenges.

In our evaluation of \emph{natural images} (also known as zero-shot scenario), we utilize CIFAR-10, CIFAR-10.1, and CIFAR-100 datasets, each comprising 10,000 images and featuring 10 and 100 classes, respectively. These datasets represent natural imagery and are novel to the model under scrutiny. Notably, CIFAR-10.1 introduces a natural domain shift from CIFAR-10, thereby enriching our assessment with varied and nuanced data distributions. We also evaluate our method on ImageNet and extend our investigation on two datasets mostly used in the field of domain generalization: PACS \cite{pacs} and OfficeHome \cite{officehome} datasets, instrumental in understanding \emph{texture and style variations}. The PACS dataset consists of 9,991 images across four domains (Art, Cartoons, Photos, Sketches) and seven classes. Lastly, the OfficeHome dataset includes 15,588 images across four domains (Art, Clipart, Product, Real) and 65 classes. Evaluating across these distinct scenarios showcases the generalizability of our method.

Transitioning to our investigation of \emph{common corruptions}, we turn to the CIFAR-10-C and CIFAR-100-C \cite{cifar10c&10.1}. These datasets offer a diverse range of 15 distinct corruptions, including elastic transform and impulse noise, among others. Each corruption is characterized by 5 severity levels, yielding a total of 75 unique testing scenarios per dataset. Within each severity level, there are 10,000 images, contributing to a comprehensive evaluation of the model's robustness against a variety of corruption types and intensities. Finally, we test our method on ImageNet-C to evaluate its performance on a larger dataset with 1,000 classes.

Finally, we examine the VisDA-C dataset's \cite{VisDA} two domain shifts: \emph{simulated} (3D) and \emph{video} (YT). The former comprises a set of 152,397 images rendered in 3D across 12 different classes. The latter includes 72,372 YouTube video frames spanning the same categories. This dataset presents an important challenge, as it bridges the gap of the type of imagery that a model can be applied on.

\section{Unsupervised clustering}

\begin{figure*}[ht!]
    \centering
    \begin{small}\setlength{\tabcolsep}{2pt}
    \begin{tabular}{cc}    
\includegraphics[width=0.5\linewidth]{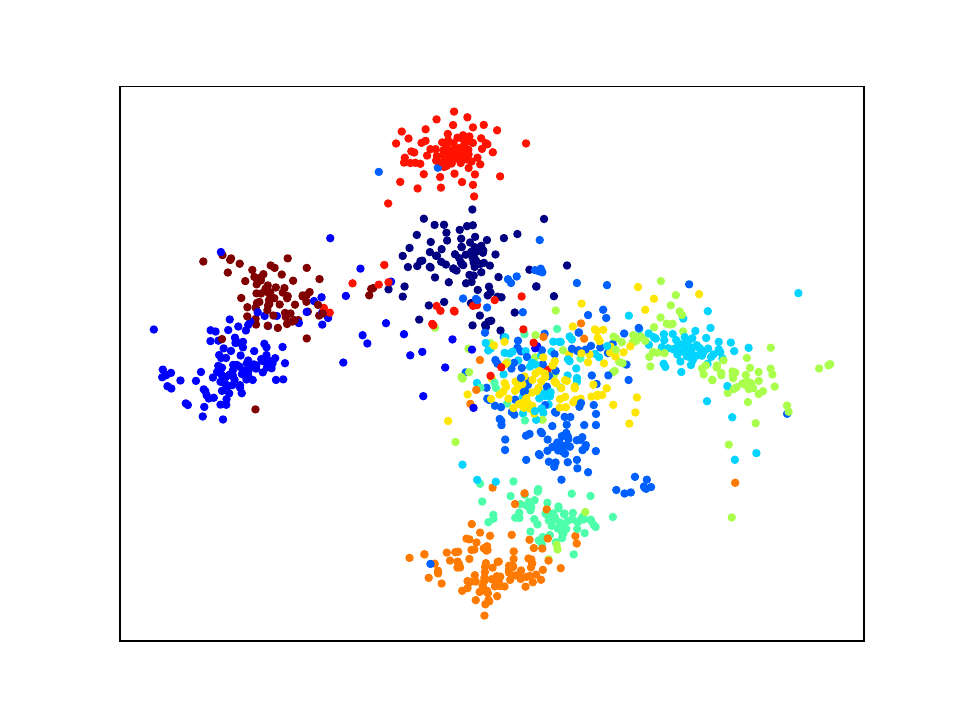} &     
    \includegraphics[width=0.5\linewidth]{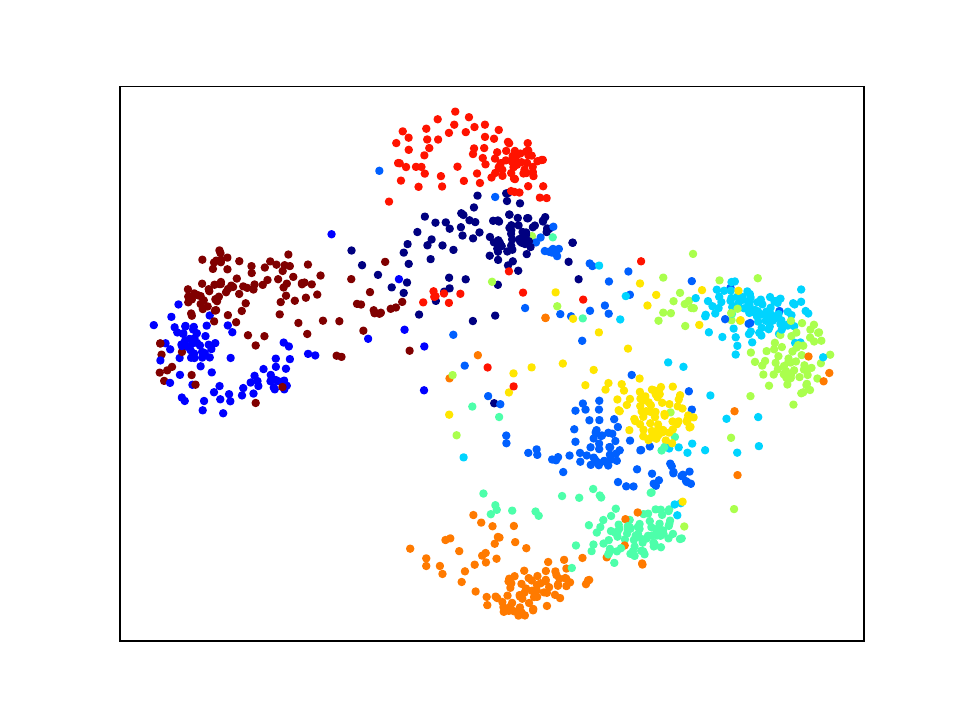}\\[-16pt]
    (a) Prediction (before adaptation) & (b) Prediction (after adaptation) \\%[2pt]
    \includegraphics[width=0.5\linewidth]{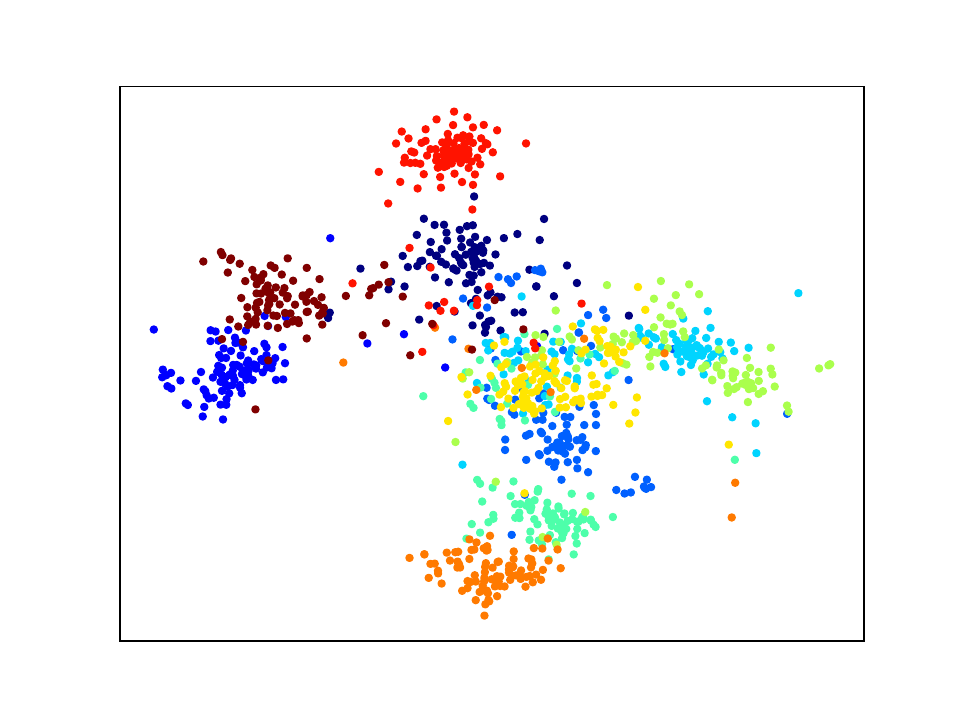} &
    \includegraphics[width=0.5\linewidth]{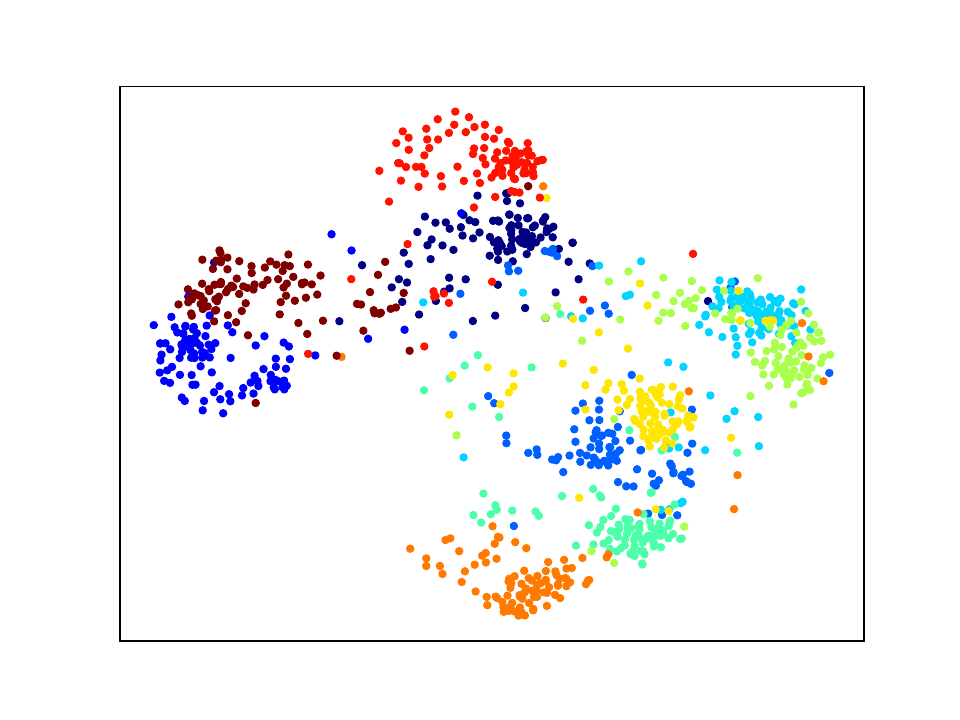}\\[-16pt]
    (c) Ground truth (before adaptation) & (d) Ground truth (after adaptation)\\[2pt]
    \end{tabular}
    \end{small}
    \caption{The t-SNE visualizations exhibit discernible attributes of brightness within the visual features derived from CLIPArTT. Panels (a) and (b) present the model's predictions before and after 10 iterations of adaptation, respectively. Panels (c) and (d) demonstrate the actual labels in the absence of adaptation and following adaptation of the representations, respectively.}
    \label{fig:tSNEcifar10c}
\end{figure*}

In Fig.~\ref{fig:tSNEcifar10c}, tSNE visualizations of data points are shown. We show how the distribution of data points change after adaptation, which improves the accuracy of class predictions and facilitates the assignment of ground truth labels.

\section{Additional settings}

We explore additional experimental settings to further explore the strengths and weaknesses of CLIPArTT. Although these scenarios deviate from the standard practices in the TTA literature, they are useful to evaluate a method's performance in potentially challenging real-world applications. For most of the following experiments, we utilize the CIFAR-10 dataset's variants, unless otherwise is mentioned.

\subsection{Imbalanced batch instances}

CLIPArTT adapts to batch data in a trandsductive manner by computing the image-to-image similarity. The diversity of the batches cannot be ensured due to their finite size. The opposite case naturally arises in large scale classification datasets such as ImageNet, where including at least one image per class in a 128$-$size batch is impossible. On the other hand, adaptation to highly imbalanced batches is understudied.

In this experiment, we force extreme imbalances in the batches, by allowing only $C$ randomly chosen classes to be present in the batch. As CLIPArTT is dependant also on the most probable predictions, it is expected that the miss-classification due to the imbalance would drive the model to directions opposite to the actual correct classes. In this experiment, we evaluate with $C = \{2, 3, 4, 5\}$ as the possible number of classes to be present in the batches, and compare CLIPArTT against TENT. Final results are shown in Table~\ref{tab:imbalance}).

It can be observed that TENT shows a higher robustness to class imbalance, as for each prediction, it depends less on the other images' predictions (i.e., induction). Interestingly, we observe a trend where TENT's performance decrease when more classes are present (e.g., $C = 5$ instead of $C = 2$), whereas CLIPArTT increase its performance as $C$ grows.

\begin{table*}[h!]
\centering
\begin{tabular}{l|cccc|cccc}
\toprule
 & \multicolumn{4}{c|}{CLIPArTT} & \multicolumn{4}{c}{TENT} \\
\midrule
 & 2 & 3 & 4 & 5 & 2 & 3 & 4 & 5 \\
\midrule
Original & 76.99 & 81.12 & 84.37 & 86.92 & 97.07 & 96.14 & 95.27 & 94.09 \\
\midrule
CIFAR-10.1 & 74.21 & 78.96 & 82.57 & 85.14 & 96.47 & 94.33 & 92.56 & 90.68 \\
\midrule
Gaussian Noise & 47.12 & 50.83 & 54.18 & 58.63 & 63.76 & 49.49 & 49.92 & 47.15 \\
Shot Noise & 48.67 & 53.30 & 56.37 & 60.24 & 69.89 & 55.83 & 56.68 & 53.55 \\
Impulse Noise & 45.98 & 49.82 & 53.15 & 57.48 & 66.87 & 54.47 & 54.77 & 52.88 \\
Defocus Blur & 60.80 & 67.78 & 71.24 & 74.85 & 89.19 & 85.39 & 83.05 & 81.44 \\
Glass Blur & 47.82 & 53.07 & 56.45 & 60.12 & 77.11 & 65.34 & 62.72 & 58.78 \\
Motion Blur & 60.85 & 66.82 & 70.67 & 73.43 & 83.33 & 79.88 & 79.28 & 75.89 \\
Zoom Blur & 56.78 & 66.52 & 70.43 & 72.94 & 86.88 & 84.15 & 81.89 & 81.35 \\
Snow & 59.20 & 66.38 & 71.29 & 73.97 & 91.84 & 87.60 & 85.32 & 82.45 \\
Frost & 60.02 & 68.08 & 72.12 & 75.15 & 92.50 & 88.62 & 86.18 & 84.25 \\
Fog & 58.57 & 66.11 & 70.54 & 73.12 & 91.26 & 87.01 & 84.07 & 83.12 \\
Brightness & 72.29 & 77.62 & 80.75 & 83.55 & 95.45 & 93.82 & 92.29 & 91.15 \\
Contrast & 62.18 & 71.68 & 75.08 & 77.64 & 89.75 & 87.11 & 85.72 & 83.25 \\
Elastic Transform & 50.85 & 58.91 & 62.12 & 65.12 & 88.17 & 82.05 & 78.13 & 76.05 \\
Pixelate & 51.75 & 57.20 & 60.42 & 63.08 & 77.35 & 71.38 & 69.22 & 66.85 \\
JPEG Compression & 49.63 & 56.08 & 59.32 & 62.05 & 79.98 & 72.34 & 69.85 & 68.17 \\
\bottomrule
\end{tabular}
\caption{Accuracy on imbalance datasets with different numbers of available classes at each batch.}
\label{tab:imbalance}
\end{table*}

\subsection{Open-Set classification}

Another interesting scenario is open-set classification, where images from classes that are not considered are present in the batch. These images could potentially affect performance, specially in transductive methods and particularly in CLIPArTT, as the image-to-image and text-to-text similarities are combined. The only available text prompts are wrongly assigned to the out-of-distribution (OOD) samples.

To measure the effectiveness of our method in this scenario, we utilize SVHN-C, a corrupted version of the SVHN~\cite{svhn} dataset, analogous to CIFAR-10.C. For each batch of 128 image, 128 additional OOD images are included. Both parts are used at the same time for adaptation, and accuracy is only measured on the first half.

As seen in Table~\ref{tab:openset}, CLIPArTT defeats TENT's open-set accuracy by a significant margin. This encouraging result suggests that our method is robust to semantic out-of-distribution perturbations.

\begin{table*}[h!]
\centering
%\dorowcolors
\small
\begin{tabular}{l|c|c}
\toprule
                  & \multicolumn{1}{c|}{CLIPArTT}    & TENT         \\ 
\midrule
Gaussian Noise    & {59.72\ppm  0.05} & 41.27\ppm 0.03  \\
Shot Noise        & {62.11\ppm  0.07} & 48.14\ppm 0.05  \\
Impulse Noise     & {55.23\ppm 0.10} & 47.86\ppm 0.07  \\
Defocus Blur      & {75.44\ppm  0.07} & 72.03\ppm 0.05  \\
Glass Blur        & {60.12\ppm  0.07} & 42.08\ppm 0.08  \\
Motion Blur       & {74.60\ppm  0.05} & 65.71\ppm 0.05  \\
Zoom Blur         & {76.15\ppm 0.05} & 71.45\ppm 0.06  \\
Snow              & {74.69\ppm  0.05} & 73.99\ppm 0.08  \\
Frost             & {77.72\ppm  0.05} & 74.49\ppm 0.09  \\
Fog               & {72.87\ppm  0.03} & 70.57\ppm 0.04  \\
Brightness        & {85.70\ppm  0.09}  & 82.34\ppm 0.05  \\
Contrast          & {76.49\ppm 0.04} & 71.67\ppm 0.04  \\
Elastic Transform & {67.42\ppm 0.01} & 65.99\ppm 0.12  \\
Pixelate          & {62.02\ppm  0.11} & 54.25\ppm 0.08  \\
JPEG Compression  & {60.81\ppm  0.07} & 54.09\ppm 0.10  \\ 
\midrule
Average  & 69.41\ppm 8.86  & 62.40\ppm 13.20 \\ 
\bottomrule
\end{tabular}
\caption{Open-set accuracy on CIFAR-10-C when using SVHN-C as the OOD dataset.}
\label{tab:openset}
\end{table*}

\subsection{Generalization after adaptation}

In TTA, the classification performance is evaluated directly on the adapted set of images in an episodic manner. However, a high accuracy on this set does not necessarily guarantee a good performance in a separate set of images unknown to the model.

Using a batch size of 160 images, we measure CLIPArTT's generalization by separating 25$\%$ of each batch as a test-time validation split. The remaining 75$\%$ is used for adaptation prior to testing on the validation split. The accuracies on the adaptation split and the validation split are both reported and contrasted against TENT. Results are shown in Table~\ref{tab:generalization}.

While TENT obtains a higher accuracy on the adaptation splits of natural images (i.e., CIFAR-10 and CIFAR-10.1), following a similar trend as in the main experiments' results, the difference in the generalization accuracy is smaller with respect to CLIPArTT's. Moreover, our method demonstrates a consistently better generalization and adaptation accuracy on corrupted samples, representing stronger domain shifts.

\begin{table*}[h!]
\centering
%\dorowcolors 
\small
\begin{tabular}{l|cc|cc}
\toprule
\multicolumn{1}{l|}{} & \multicolumn{2}{c|}{CLIPArTT} & \multicolumn{2}{c}{TENT} \\ \cmidrule{2-5} 
\multicolumn{1}{l|}{} & Acc.              & Acc. Gen.          & Acc.            & Acc. Gen.       \\ \midrule
Original             & 95.43\ppm 0.01       & 71.23\ppm 0.0008      & 97.02\ppm 0.02     & 72.68\ppm 0.0023   \\ \midrule
CIFAR-10.1           & 88.75\ppm 0.85       & 65.87\ppm 0.8083      & 90.49\ppm 0.38     & 67.07\ppm 1.1      \\ \midrule
Gaussian Noise       & 63.79\ppm 0.20        & 46.87\ppm 0.0008      & 45.30\ppm 0.05     & 30.83\ppm 0.0030   \\
Shot Noise           & 66.38\ppm 0.10        & 79.80\ppm 0.0038      & 50.65\ppm 0.18     & 35.85\ppm 0.0025   \\
Impulse Noise        & 59.31\ppm 0.19       & 44.56\ppm 0.0023      & 51.91\ppm 0.20     & 37.99\ppm 0.0037   \\
Defocus Blur         & 81.26\ppm 0.03       & 60.13\ppm 0.0016      & 84.48\ppm 0.07     & 60.32\ppm 0.0028   \\
Glass Blur           & 65.38\ppm 0.04       & 49.47\ppm 0.0020      & 56.68\ppm 0.26     & 41.32\ppm 0.0035   \\
Motion Blur          & 80.41\ppm 0.12       & 60.08\ppm 0.0022      & 75.70\ppm 0.13     & 55.72\ppm 0.0013   \\
Zoom Blur            & 81.97\ppm 0.05       & 60.63\ppm 0.0009      & 81.14\ppm 0.13     & 59.33\ppm 0.0007   \\
Snow                 & 82.06\ppm 0.14       & 60.83\ppm 0.0014      & 83.08\ppm 0.13     & 61.08\ppm 0.0008   \\
Frost                & 83.85\ppm 0.19       & 61.73\ppm 0.0009      & 84.70\ppm 0.11     & 66.75\ppm 0.0006   \\
Fog                  & 90.04\ppm 0.27       & 59.48\ppm 0.0027      & 81.56\ppm 0.09     & 60.69\ppm 0.0033   \\
Brightness           & 91.73\ppm 0.04       & 68.01\ppm 0.0017      & 92.93\ppm 0.03     & 68.80\ppm 0.0018   \\
Contrast             & 82.63\ppm 0.07       & 61.28\ppm 0.0025      & 83.53\ppm 0.06     & 62.97\ppm 0.0009   \\
Elastic Transform    & 74.04\ppm 0.22       & 54.87\ppm 0.0011      & 74.08\ppm 0.10     & 54.92\ppm 0.0009   \\
Pixelate             & 70.58\ppm 0.02        & 51.53\ppm 0.0042      & 67.35\ppm 0.18     & 50.19\ppm 0.0004   \\
JPEG Compression     & 67.32\ppm 0.06       & 49.56\ppm 0.0023      & 66.25\ppm 0.14     & 48.85\ppm 0.0021   \\ \midrule
Average              & \textbf{75.38}\ppm 9.04      & 55.92\ppm 6.62        & 71.78\ppm 14.14    & 52.77\ppm 11.10    \\ \bottomrule
\end{tabular}
\caption{Generalization results on CIFAR-10 variants. Each batch is divided into an adaptation split and a validation split. Accuracy (Acc.) is measured on the former after adaptation, whilst the generalization accuracy (Acc. Gen.) is measured on the later.}
\label{tab:generalization}
\end{table*}

\section{Computational Cost}
In this section, we compare the computational cost of CLIPArTT with other TTA methods through a thorough evaluation under consistent conditions, using an NVIDIA A6000 GPU within the same Python environment. The provided table \ref{tab:computationalcost} compares adaptation time, memory usage, and the number of learnable parameters across various TTA methods, including our proposed CLIPArTT. The results demonstrate that CLIPArTT maintains competitive adaptation time and memory usage relative to other approaches, such as TENT and TPT.

\begin{table*}[h!]
    \centering 
    \small
    \begin{tabular}{l|c|c|c}
        \toprule
        Method & \makecell[c]{Adaptation \\ Time} & Memory & \makecell[c]{Pct. of Learnable \\ Parameters} \\
        \midrule
        TENT   & 0.28 s & 1.5 GB & 0.026\% \\ 
        TPT & 0.26 s & 1.7 GB & 0.001\% \\
        CLIPArTT & 0.55 s & 1.7 GB & 0.026\% \\
        \bottomrule
    \end{tabular}
    \caption{Comparison of Computational Cost.}
    \label{tab:computationalcost}
\end{table*}

\section{Comprehensive experimental results}

We present comprehensive tables containing all the detailed information about results that was summarized in the main paper.

\begin{table}[!h]
    \centering    
    \dorowcolors    
    \begin{small}
    \setlength{\tabcolsep}{5pt}
    \begin{tabular}{l|cc|cc}
    \toprule
        % ~ & CIFAR10 & ~ & CIFAR100 & ~ \\ \midrule
        ~ & \multicolumn{2}{c|}{CIFAR10} & \multicolumn{2}{c}{CIFAR100} \\ \cmidrule(lr){2-3} \cmidrule(lr){4-5}
        ~ & Top 1 & Top 3 & Top 1 & Top 3 \\ \midrule
        ORIGINAL & 88.74 & 100.00 & 61.68 & 97.34 \\  \midrule
        Gaussian Noise & 35.27 & 99.87 & 14.8 & 63.66 \\ 
        Shot noise & 39.67 & 99.99 & 16.03 & 67.02 \\ 
        Impulse Noise & 42.61 & 100.00 & 13.85 & 64.4 \\ 
        Defocus blur & 69.76 & 100.00 & 36.74 & 90.14 \\ 
        Glass blur & 42.40 & 100.00 & 14.19 & 61.66 \\ 
        Motion blur & 63.97 & 100.00 & 36.14 & 90.36 \\ 
        Zoom blur & 69.83 & 100.00 & 40.24 & 91.27 \\ 
        Snow & 71.78 & 100.00 & 38.95 & 91.40 \\ 
        Frost & 72.86 & 100.00 & 40.56 & 92.23 \\ 
        Fog & 67.04 & 99.98 & 38.00 & 91.51 \\ 
        Brightness & 81.87 & 100.00 & 48.18 & 93.10 \\ 
        Contrast & 64.37 & 100.00 & 29.53 & 84.67 \\ 
        Elastic transform & 60.83 & 100.00 & 26.33 & 78.96 \\ 
        Pixelate & 50.53 & 100.00 & 21.98 & 75.65 \\ 
        JPEG compression & 55.48 & 100.00 & 25.91 & 80.81 \\ \midrule
        Average & 59.22 & 99.99 & 29.43 & 81.12 \\ \bottomrule
    \end{tabular}
    \end{small}
    \caption{Accuracy (\%) on CIFAR-10/100 and CIFAR-10/100-C datasets with Level 5 corruption for the top 1 or the top 3 predicted classes.}
	\label{tab:MotivationPrediction2}
\end{table}

\begin{table}[!ht]
    \centering
    \dorowcolors
    \setlength{\tabcolsep}{5pt}
    \small
    \begin{tabular}{l|ccc}
    \toprule
        & K = 1 & K = 3 & K = 4 \\ \midrule
        ORIGINAL & 89.8\ppm0.05 & 90.04\ppm0.13 & 90.41\ppm0.07 \\ \midrule
        CIFAR 10.1 & 85.37\ppm0.17 & 86.35\ppm0.27 & 86.07\ppm0.21 \\ \midrule
        Gaussian Noise & 60.2\ppm0.24 & 59.90\ppm0.36 & 59.71\ppm0.15 \\ 
        Shot noise & 62.08\ppm0.11 & 62.77\ppm0.07 & 62.17\ppm0.16 \\ 
        Impulse Noise & 54.33\ppm0.07 & 56.02\ppm0.16 & 56.27\ppm0.15 \\ 
        Defocus blur & 77.16\ppm0.02 & 76.74\ppm0.05 & 76.79\ppm0.11 \\ 
        Glass blur & 61.91\ppm0.15 & 61.77\ppm0.16 & 61.72\ppm0.23 \\ 
        Motion blur & 74.94\ppm0.15 & 76.01\ppm0.19 & 76.33\ppm0.1 \\ 
        Zoom blur & 76.84\ppm0.13 & 77.40\ppm0.20 & 77.15\ppm0.04 \\ 
        Snow & 76.87\ppm0.05 & 77.29\ppm0.16 & 76.56\ppm0.16 \\ 
        Frost & 77.81\ppm0.04 & 79.20\ppm0.08 & 78.42\ppm0.04 \\ 
        Fog & 75.83\ppm0.28 & 75.74\ppm0.14 & 75.65\ppm0.06 \\ 
        Brightness & 85.55\ppm0.12 & 86.59\ppm0.16 & 86.83\ppm0.1 \\ 
        Contrast & 78.02\ppm0.18 & 77.82\ppm0.14 & 78.27\ppm0.14 \\ 
        Elastic transform & 69.42\ppm0.07 & 70.20\ppm0.01 & 69.81\ppm0.2 \\ 
        Pixelate & 66.07\ppm0.09 & 66.52\ppm0.13 & 66.45\ppm0.08 \\ 
        JPEG compression & 64.82\ppm0.26 & 63.51\ppm0.14 & 62.72\ppm0.25 \\ \midrule
        Average & 70.79 & 71.17 & 70.99 \\ \bottomrule
    \end{tabular}
    \caption{Accuracy (\%) on CIFAR-10, CIFAR-10.1 and CIFAR-10-C datasets with Level 5 corruption for different number of K selected classes to create pseudo-label.}
	\label{tab:KCifar102}
\end{table}

\begin{table}[!h]
    \centering
    \begin{footnotesize}
    \dorowcolors
    \resizebox{\textwidth}{!}{
    \setlength{\tabcolsep}{5pt}
    \begin{tabular}{l|cccccc}
    \toprule
        & CLIP & BS = 8 & BS = 16 & BS = 32 & BS = 64 & BS = 128 \\ \midrule
        ORIGINAL & 88.74 & 82.50\ppm0.13 & 85.89\ppm0.19 & 88.25\ppm0.15 & 89.48\ppm0.15 & 90.04\ppm0.13 \\ \midrule
        CIFAR 10.1 & 83.25 & 77.2\ppm0.92 & 81.55\ppm0.53 & 84.00\ppm0.31 & 85.40\ppm0.08 & 86.35\ppm0.27 \\ \midrule
        Gaussian Noise & 35.27 & 47.30\ppm0.37 & 50.91\ppm0.35 & 54.23\ppm0.28 & 57.89\ppm0.13 & 59.90\ppm0.36 \\ 
        Shot noise & 39.67 & 49.62\ppm0.26 & 53.1\ppm0.27 & 56.88\ppm0.23 & 60.56\ppm0.12 & 62.77\ppm0.07 \\ 
        Impulse Noise & 42.61 & 47.24\ppm0.22 & 50.24\ppm0.48 & 52.7\ppm0.21 & 54.88\ppm0.17 & 56.02\ppm0.16 \\ 
        Defocus blur & 69.76 & 68.24\ppm0.35 & 72.22\ppm0.04 & 75.09\ppm0.16 & 75.97\ppm0.27 & 76.74\ppm0.05 \\ 
        Glass blur & 42.40 & 49.49\ppm0.30 & 53.27\ppm0.04 & 57.18\ppm0.24 & 60.12\ppm0.14 & 61.77\ppm0.16 \\ 
        Motion blur & 63.97 & 65.22\ppm0.06 & 69.02\ppm0.30 & 72.54\ppm0.27 & 74.71\ppm0.18 & 76.01\ppm0.19 \\ 
        Zoom blur & 69.83 & 67.69\ppm0.20 & 71.33\ppm0.11 & 74.53\ppm0.11 & 76.35\ppm0.07 & 77.40\ppm0.20 \\ 
        Snow & 71.78 & 68.68\ppm0.42 & 72.37\ppm0.11 & 74.93\ppm0.18 & 76.53\ppm0.41 & 77.29\ppm0.16 \\ 
        Frost & 72.86 & 70.35\ppm0.25 & 73.93\ppm0.34 & 76.81\ppm0.23 & 78.22\ppm0.13 & 79.20\ppm0.08 \\ 
        Fog & 67.04 & 66.25\ppm0.31 & 69.71\ppm0.24 & 72.36\ppm0.23 & 73.96\ppm0.21 & 75.74\ppm0.14 \\ 
        Brightness & 81.87 & 77.36\ppm0.17 & 81.20\ppm0.20 & 84.07\ppm0.08 & 85.58\ppm0.25 & 86.59\ppm0.16 \\ 
        Contrast & 64.37 & 65.12\ppm0.07 & 69.02\ppm0.12 & 72.60\ppm0.46 & 75.79\ppm0.24 & 77.82\ppm0.14 \\ 
        Elastic transform & 60.83 & 59.61\ppm0.11 & 63.67\ppm0.13 & 66.36\ppm0.26 & 68.74\ppm0.07 & 70.20\ppm0.01 \\ 
        Pixelate & 50.53 & 56.78\ppm0.24 & 60.01\ppm0.06 & 62.57\ppm0.19 & 64.64\ppm0.03 & 66.52\ppm0.13 \\ 
        JPEG compression & 55.48 & 57.59\ppm0.26 & 60.78\ppm0.12 & 62.63\ppm0.06 & 63.43\ppm0.16 & 63.51\ppm0.14 \\ \midrule
        Average & 59.22 & 61.10 & 64.72 & 67.70 & 69.82 & 71.17 \\ \bottomrule
    \end{tabular}}
    \end{footnotesize}
    \caption{Accuracy (\%) on CIFAR-10, CIFAR-10.1 and CIFAR-10-C datasets with Level 5 corruption for different batch sizes.}
	\label{tab:BSCIfar102}
\end{table}

\begin{table}[!ht]
    \centering
    \dorowcolors
    \small
    \begin{tabular}{l|cccc}    
    \toprule
        & CLIP & TENT & TPT\,{\scriptsize(BS=32)} & CLIPArTT \\ \midrule
        \textsc{original} & 89.25 & \textbf{92.75\ppm0.17} & 89.80\ppm0.05 & 92.61\ppm0.05 \\ \midrule
        CIFAR 10.1 & 84.00 & 88.52\ppm0.33 & 83.75.0.21\ppm & \textbf{88.72\ppm0.33} \\ \midrule
        Gaussian Noise & 37.75 & 31.04\ppm0.38 & 35.35\ppm0.15 & \textbf{60.89\ppm0.26} \\ 
        Shot noise & 41.10 & 40.54\ppm0.41 & 41.03\ppm0.19 & \textbf{65.19\ppm0.21} \\ 
        Impulse Noise & 51.71 & 58.03\ppm0.16 & 54.86\ppm0.07 & \textbf{67.55\ppm0.09} \\ 
        Defocus blur & 70.07 & 77.57\ppm0.03 & 70.29\ppm0.02 & \textbf{78.92\ppm0.12} \\ 
        Glass blur & 42.24 & 47.16\ppm0.05 & 37.86\ppm0.17 & \textbf{57.18\ppm0.20} \\ 
        Motion blur & 65.81 & 76.16\ppm0.05 & 67.43\ppm0.11 & \textbf{76.59\ppm0.06} \\ 
        Zoom blur & 72.50 & \textbf{79.64\ppm0.12} & 72.91\ppm0.02 & 79.62\ppm0.11 \\ 
        Snow & 73.23 & \textbf{81.68\ppm0.03} & 72.98\ppm0.32 & 81.13\ppm0.29 \\ 
        Frost & 76.52 & \textbf{83.22\ppm0.05} & 75.87\ppm0.16 & 81.24\ppm0.08 \\ 
        Fog & 68.35 & \textbf{80.78\ppm0.15} & 69.13\ppm0.27 & 78.47\ppm0.19 \\ 
        Brightness & 83.36 & \textbf{89.85\ppm0.11} & 83.67\ppm0.14 & 88.66\ppm0.15 \\ 
        Contrast & 61.90 & \textbf{79.24\ppm0.19} & 62.16\ppm0.06 & 75.15\ppm0.07 \\ 
        Elastic transform & 53.16 & 62.54\ppm0.08 & 51.26\ppm0.23 & \textbf{69.49\ppm0.08} \\ 
        Pixelate & 48.48 & 67.08\ppm0.24 & 44.65\ppm0.21 & \textbf{71.80\ppm0.16} \\ 
        JPEG compression & 56.05 & 65.42\ppm0.05 & 56.73\ppm0.07 & \textbf{66.42\ppm0.25} \\ \midrule
        Average & 60.15 & 68.00 & 59.75 & \textbf{73.22} \\ \bottomrule
    \end{tabular}
    \caption{Accuracy (\%) on CIFAR-10, CIFAR-10.1 and CIFAR-10-C datasets with ViT-B/16 as visual encoder.}
	\label{tab:B16Cifar10}
\end{table}

\begin{table}[!ht]
    \centering
    \dorowcolors
    \small
    \setlength{\tabcolsep}{5pt}
    \begin{tabular}{l|cccc}
    \toprule
        & CLIP & TENT & TPT\,{\scriptsize(BS=32)} & CLIPArTT \\ \midrule
        \textsc{original} & 95.36 & \textbf{96.13\ppm0.06} & 95.18\ppm0.02 & 95.16\ppm0.03 \\ \midrule
        CIFAR 10.1 & 91.20 & \textbf{92.22\ppm0.25} & 91.32\ppm0.12 & 91.02\ppm0.02 \\ \midrule
        Gaussian Noise & 64.64 & 68.87\ppm0.20 & 64.44\ppm0.11 & \textbf{70.04\ppm0.31} \\ 
        Shot noise & 67.82 & \textbf{71.95\ppm0.06} & 66.81\ppm0.19 & 71.44\ppm0.16 \\ 
        Impulse Noise & 78.21 & \textbf{80.22\ppm0.19} & 76.46\ppm0.17 & 79.42\ppm0.15 \\ 
        Defocus blur & 80.73 & \textbf{83.10\ppm0.03} & 79.01\ppm0.23 & 81.75\ppm0.19 \\ 
        Glass blur & 50.29 & 57.12\ppm0.07 & 49.64\ppm0.23 & \textbf{58.13\ppm0.23} \\ 
        Motion blur & 80.75 & \textbf{82.69\ppm0.11} & 78.85\ppm0.04 & 80.76\ppm0.12 \\ 
        Zoom blur & 82.75 & \textbf{84.91\ppm0.08} & 82.32\ppm0.13 & 83.39\ppm0.05 \\ 
        Snow & 83.01 & \textbf{85.99\ppm0.11} & 82.69\ppm0.10 & 84.48\ppm0.07 \\ 
        Frost & 84.90 & \textbf{87.15\ppm0.12} & 84.63\ppm0.08 & 85.21\ppm0.06 \\ 
        Fog & 78.44 & \textbf{81.30\ppm0.07} & 77.56\ppm0.17 & 79.27\ppm0.07 \\ 
        Brightness & 91.67 & \textbf{93.07\ppm0.04} & 90.94\ppm0.04 & 91.87\ppm0.09 \\ 
        Contrast & 84.20 & \textbf{87.93\ppm0.04} & 82.88\ppm0.09 & 86.19\ppm0.06 \\ 
        Elastic transform & 65.45 & \textbf{69.96\ppm0.12} & 64.81\ppm0.14 & 67.43\ppm0.24 \\ 
        Pixelate & 75.10 & \textbf{77.89\ppm0.05} & 72.92\ppm0.12 & 77.11\ppm0.10 \\ 
        JPEG compression & 72.58 & \textbf{75.49\ppm0.07} & 71.18\ppm0.19 & 74.46\ppm0.11 \\ \midrule
        Average & 76.04 & \textbf{79.18} & 75.01 & 78.06 \\ \bottomrule
    \end{tabular}
    \caption{Accuracy (\%) on CIFAR-10, CIFAR-10.1 and CIFAR-10-C datasets with ViT-L/14 as visual encoder.}
	\label{tab:L14Cifar10}
\end{table}

\begin{table}[!ht]
    \centering
    \dorowcolors
    \setlength{\tabcolsep}{5pt}
    \small
    \begin{tabular}{l|cccc}
    \toprule
        & CLIP & TENT & TPT\,{\scriptsize(BS=32)} & CLIPArTT \\ \midrule
        \textsc{original} & 64.76 & \textbf{71.73\ppm0.14} & 67.15\ppm0.23 & 71.34\ppm0.07 \\  \midrule
        Gaussian Noise & 15.88 & 12.28\ppm0.20 & 15.43\ppm0.03 & \textbf{19.01\ppm0.24} \\ 
        Shot noise & 17.49 & 15.07\ppm0.21 & 16.88\ppm0.07 & \textbf{20.27\ppm0.21} \\ 
        Impulse Noise & 21.43 & 13.13\ppm0.16 & \textbf{22.12\ppm0.15} & 17.66\ppm0.10 \\ 
        Defocus blur & 40.10 & \textbf{50.35\ppm0.03} & 41.08\ppm0.22 & 49.86\ppm0.13 \\ 
        Glass blur & 13.48 & \phz{}4.84\ppm0.14 & \textbf{18.43\ppm0.15} & 18.34\ppm0.31 \\ 
        Motion blur & 39.82 & 49.85\ppm0.37 & 40.85\ppm0.26 & \textbf{50.00\ppm0.09} \\ 
        Zoom blur & 45.45 & \textbf{54.76\ppm0.04} & 46.77\ppm0.06 & 54.13\ppm0.08 \\ 
        Snow & 42.77 & 52.38\ppm0.18 & 47.24\ppm0.18 & \textbf{52.80\ppm0.27} \\ 
        Frost & 45.39 & \textbf{51.66\ppm0.04} & 48.61\ppm0.14 & 49.56\ppm0.08 \\ 
        Fog & 38.98 & \textbf{50.74\ppm0.14} & 39.92\ppm0.16 & 49.92\ppm0.11 \\ 
        Brightness & 52.55 & \textbf{64.26\ppm0.09} & 55.83\ppm0.10 & 63.76\ppm0.13 \\ 
        Contrast & 33.32 & \textbf{48.69\ppm0.08} & 33.13\ppm0.16 & 47.86\ppm0.02 \\ 
        Elastic transform & 24.39 & \textbf{33.56\ppm0.28} & 27.36\ppm0.10 & 32.93\ppm0.23 \\ 
        Pixelate & 21.89 & 36.20\ppm0.28 & 21.26\ppm0.10 & \textbf{39.49\ppm0.21} \\ 
        JPEG compression & 27.21 & 30.80\ppm0.05 & 30.97\ppm0.10 & \textbf{35.56\ppm0.23} \\  \midrule
        Average & 32.01 & 37.90 & 33.73 & \textbf{40.08} \\ \bottomrule
    \end{tabular}
    \caption{Accuracy (\%) on CIFAR-100 and CIFAR-100-C datasets with ViT-B/16 as visual encoder.}
	\label{tab:B16Cifar100}
\end{table}

\begin{table}[!ht]
    \centering
    \dorowcolors
    \small
    \setlength{\tabcolsep}{5pt}
    \begin{tabular}{l|cccc}
    \toprule
        & CLIP & TENT & TPT\,{\scriptsize(BS=16)} & CLIPArTT \\ \midrule
        ORIGINAL & 73.28 & 78.03\ppm0.08 & 76.85\ppm0.06 & \textbf{79.42\ppm0.08} \\ \midrule
        Gaussian Noise & 30.55 & 36.93\ppm0.03 & 36.10\ppm0.11 & \textbf{41.46\ppm0.15} \\ 
        Shot noise & 34.58 & 40.96\ppm0.16 & 38.23\ppm0.13 & \textbf{44.27\ppm0.09} \\ 
        Impulse Noise & 44.89 & 49.09\ppm0.14 & 49.69\ppm0.21 & \textbf{51.44\ppm0.23} \\ 
        Defocus blur & 48.88 & 55.23\ppm0.07 & 50.43\ppm0.19 & \textbf{56.55\ppm0.22} \\ 
        Glass blur & 23.46 & 27.02\ppm0.23 & 24.35\ppm0.22 & \textbf{30.47\ppm0.14} \\ 
        Motion blur & 50.83 & 56.03\ppm0.20 & 51.94\ppm0.04 & \textbf{56.98\ppm0.18} \\ 
        Zoom blur & 56.02 & 61.19\ppm0.10 & 56.96\ppm0.16 & \textbf{62.56\ppm0.04} \\ 
        Snow & 49.03 & 55.60\ppm0.09 & 54.89\ppm0.11 & \textbf{58.81\ppm0.11} \\ 
        Frost & 53.27 & 58.21\ppm0.15 & 58.15\ppm0.33 & \textbf{60.38\ppm0.23} \\ 
        Fog & 48.51 & 53.37\ppm0.25 & 49.26\ppm0.13 & \textbf{54.38\ppm0.04} \\ 
        Brightness & 60.53 & 67.34\ppm0.17 & 66.60\ppm0.10 & \textbf{69.63\ppm0.14} \\ 
        Contrast & 50.24 & 59.91\ppm0.13 & 53.64\ppm0.24 & \textbf{63.39\ppm0.13} \\ 
        Elastic transform & 35.07 & 38.49\ppm0.12 & 35.72\ppm0.09 & \textbf{39.57\ppm0.39} \\ 
        Pixelate & 43.86 & 48.37\ppm0.17 & 44.32\ppm0.10 & \textbf{50.45\ppm0.16} \\ 
        JPEG compression & 39.11 & 44.42\ppm0.09 & 43.44\ppm0.11 & \textbf{47.45\ppm0.14} \\ \midrule
        Average & 44.59 & 50.14 & 47.58 & \textbf{52.52} \\ \bottomrule
    \end{tabular}
    \caption{Accuracy (\%) on CIFAR-100 and CIFAR-100-C datasets with ViT-L/14 as visual encoder.}
	\label{tab:L14Cifar100}
\end{table}

\bibliographystyle{unsrtnat}
\bibliography{main} 

\end{document}

%% file: preamble.tex
\usepackage{times}
\usepackage{epsfig}
\usepackage{graphicx}
\usepackage{amsmath}
\usepackage{amssymb}
\usepackage{amsthm}
\usepackage[table]{xcolor}
\usepackage{multirow}
\usepackage{booktabs}
\usepackage[accsupp]{axessibility}
\usepackage{bm}
\usepackage{xfrac}
\usepackage{makecell}
\usepackage{hyperref}
\usepackage{natbib}
\usepackage{verbatim}
\usepackage{bbm}

\newcommand{\mypar}[1]{\vspace{3pt}
\noindent\textbf{#1~}}
\newcommand{\tr}[1]{{#1^{\top}}}
\newcommand{\mr}[1]{\mathrm{#1}}

\newcommand{\ppm}{\,\scriptsize$\pm$}

\newcommand{\zz}{\mathbf{z}}
\newcommand{\xx}{\mathbf{x}}

\newcommand{\yy}{\mathbf{y}}
\newcommand{\ttt}{\mathbf{t}}

\newcommand{\LL}{\mathbf{L}}
\newcommand{\II}{\mathbf{I}}

\newcommand{\ZZ}{\mathbf{Z}}
\newcommand{\PP}{\mathbf{P}}
\newcommand{\QQ}{\mathbf{Q}}

\newcommand{\WW}{\mathbf{W}}
\newcommand{\softmax}{\text{softmax}}

\newcommand{\R}{\real}
\newcommand{\Sim}{\mathbf{S}}

\newcommand{\loss}{\mathcal{L}}
\newcommand{\real}{\mathbb{R}}

\newcommand{\cmark}{\ding{51}}%
\newcommand{\xmark}{\ding{55}}

\newcommand{\dorowcolors}{\rowcolors{2}{gray!10}{white}}
\newcommand{\phz}{\phantom{0}}

\usepackage{pifont}
\usepackage{graphicx,scalerel}

\definecolor{myred}{rgb}{1, 0, 0}
\definecolor{mygreen}{rgb}{0, .67, 0}